\documentclass[final]{cvpr}

\usepackage{times}
\usepackage{epsfig}
\usepackage{graphicx}
\usepackage{amsmath}
\usepackage{amssymb}
\usepackage{caption}
\usepackage{subcaption}

\usepackage[pagebackref=true,breaklinks=true,colorlinks,bookmarks=false]{hyperref}

\def\cvprPaperID{12} 
\def\confYear{CVPR 2021}
\begin{document}

\title{Instagram Filter Removal on Fashionable Images}

\author{Furkan Kınlı\textsuperscript{1} \qquad Barış Özcan\textsuperscript{2} \qquad Furkan Kıraç\textsuperscript{3} \\
Video, Vision and Graphics Lab\\
Özyeğin University\\
{\tt\small \{furkan.kinli\textsuperscript{1}, furkan.kirac\textsuperscript{3}\}@ozyegin.edu.tr, baris.ozcan.10097@ozu.edu.tr\textsuperscript{2}}

}
\maketitle

\begin{abstract}
   Social media images are generally transformed by filtering to obtain aesthetically more pleasing appearances. However, CNNs generally fail to interpret both the image and its filtered version as the same in the visual analysis of social media images. We introduce Instagram Filter Removal Network (IFRNet) to mitigate the effects of image filters for social media analysis applications. To achieve this, we assume any filter applied to an image substantially injects a piece of additional style information to it, and we consider this problem as a reverse style transfer problem. The visual effects of filtering can be directly removed by adaptively normalizing external style information in each level of the encoder. Experiments demonstrate that IFRNet outperforms all compared methods in quantitative and qualitative comparisons, and has the ability to remove the visual effects to a great extent. Additionally, we present the filter classification performance of our proposed model, and analyze the dominant color estimation on the images unfiltered by all compared methods.
\end{abstract}

\section{Introduction}

In recent years, hundred of millions of photos have been shared on different social media platforms (\eg Instagram, Facebook, etc.). Predicting the preferences of the users by analyzing their posts on such platforms becomes a crucial element of increasing the profitability for different industries (\eg food, e-commerce and fashion). This analysis is roughly composed of understanding contents of a bunch of images and extracting particular information that may be useful for a certain domain. The first part of this composition, \textit{visual understanding}, is mostly attacked by various deep learning approaches and their recent developments \cite{Guo2016DeepLF}.

\begin{figure}[t!]
     \centering
     \begin{subfigure}{\linewidth}
         \includegraphics[width=0.322\linewidth]{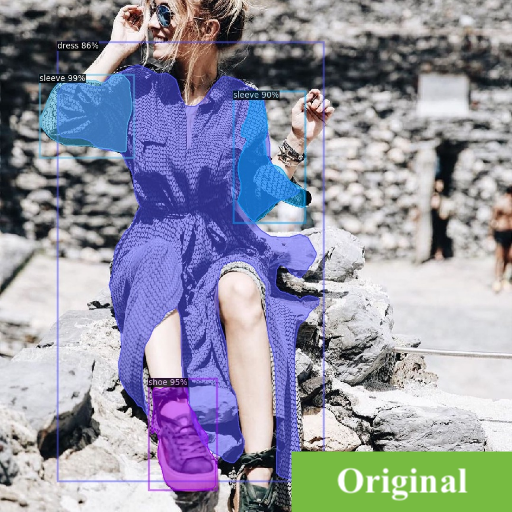}
         \includegraphics[width=0.322\linewidth]{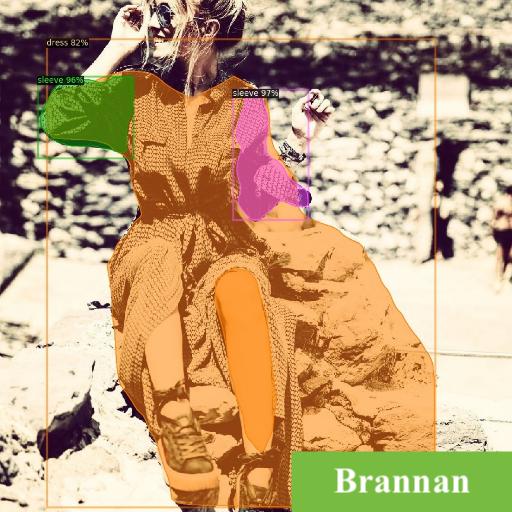}
         \includegraphics[width=0.322\linewidth]{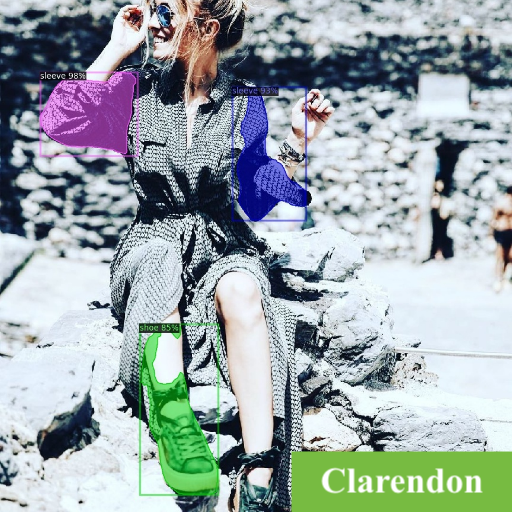}
     \end{subfigure}
     \begin{subfigure}{\linewidth}
         \includegraphics[width=0.322\linewidth]{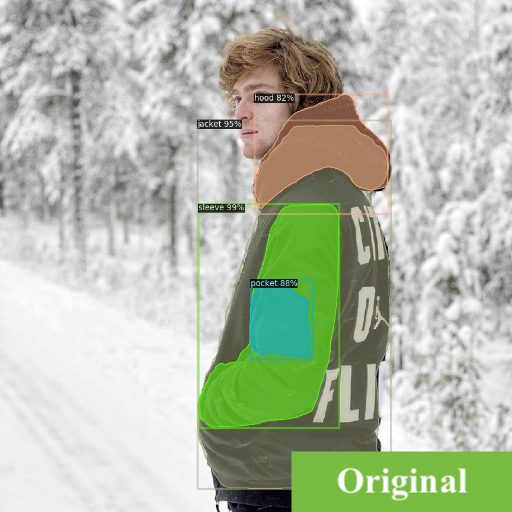}
         \includegraphics[width=0.322\linewidth]{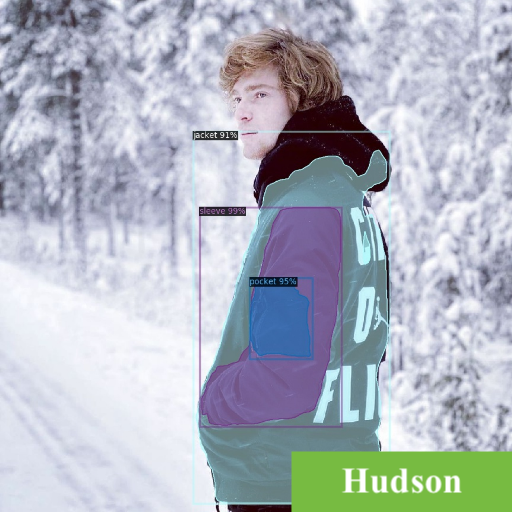}
         \includegraphics[width=0.322\linewidth]{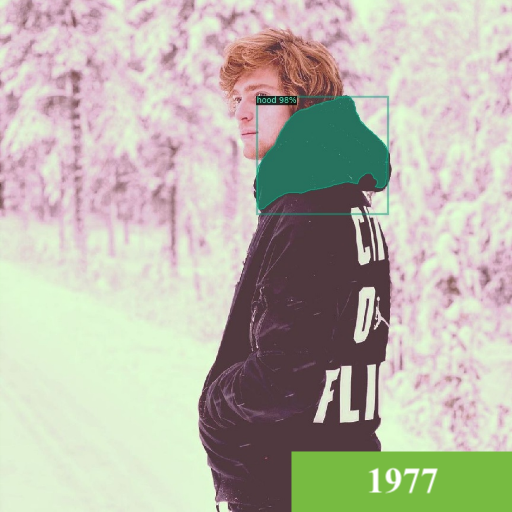}
     \end{subfigure}
     \begin{subfigure}{\linewidth}
         \includegraphics[width=0.322\linewidth]{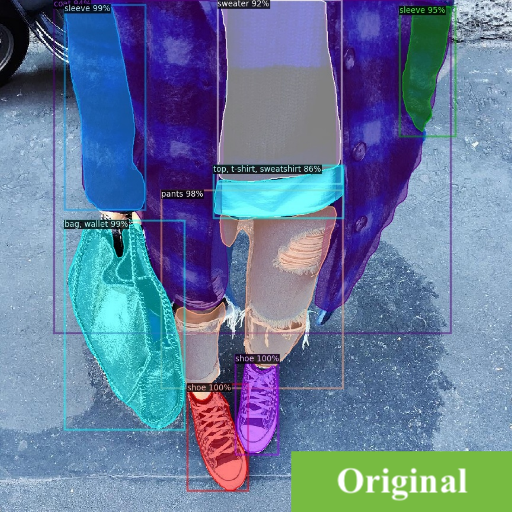}
         \includegraphics[width=0.322\linewidth]{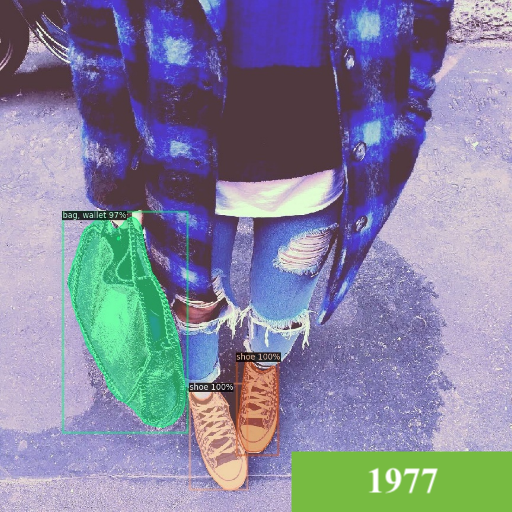}
         \includegraphics[width=0.322\linewidth]{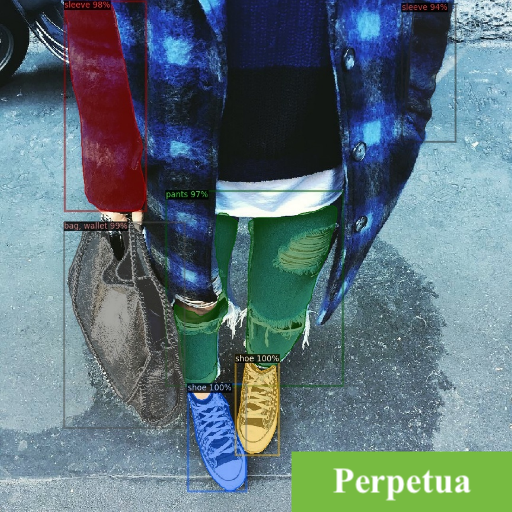}
     \end{subfigure}
        \caption{Failure cases arising from filtering on detection and segmentation tasks in fashion domain. Example images are predicted by Attr-Mask-RCNN trained on Fashionpedia dataset, which is introduced in \cite{Jia2020Fashionpedia}.}
        \label{fig:fig1}
\end{figure}

Visual understanding for a specific purpose may contain several different tasks to be accomplished, such as image classification, object detection, instance segmentation and image retrieval. Convolutional Neural Networks (CNNs) show an outstanding performance on visual understanding tasks, and there are several prominent studies \cite{8237584, He2015, triplet, Koch2015SiameseNN, Redmon_2016_CVPR, NIPS2015_14bfa6bb, Simonyan15} proposing different fundamental solutions to these tasks by employing the variants of CNNs. However, CNNs may not deliver the same performance in real-world applications, as in the standard benchmark studies, due to the varied distractive factors like noise or blurring in real-world images or different transformations applied to the images. Recent studies \cite{artisticFilter, hendrycks2018benchmarking, 45818, Wu_Wu_Singh_Davis_2020} have shown that CNNs are sensitive to these distractive factors, and they lead to degrade the performance significantly.

With the intent of exhibiting aesthetically more pleasing appearances or scenes, the images shared on social media platforms are mostly transformed into a different version by applying some filters. These filters modify the original image in many different ways (\eg adjusting contrast, brightness, hue, saturation; introducing different levels of blur and noise; applying color curves or vignetting). These modifications not only make an image more aesthetically pleasing, but also interpolate a particular style information to its feature maps. Considering the performance of visual understanding systems, it is important to handle such interpolations without turning the problem into an ill-posed problem. Note that detecting the levels of modifications applied during application of the filters for each single image is an ill-posed problem. At this point, CNNs do not give the exact outputs for an image and its filtered version, as shown in Figure \ref{fig:fig1}, due to the differences on their feature maps caused by from filtering. This leads to the performance degradation on  visual understanding tasks in various domains.

Previous studies addressing this issue have proposed different solutions that try to classify the specific filter applied to the images \cite{bianco2017artistic, filterInvariant, chu2019photo, Wu_Wu_Singh_Davis_2020} or to learn the complex parameters of a set of transformations applied to the images \cite{artisticFilter, photoTransform}. Although these solutions have the ability to predict the filter or a set of transformations applied, they could not recover the original image. In this study, we assume that any filter applied to an image basically stands for the additional style information injected to the images. Therefore, we consider this problem as a \textit{reverse} style transfer problem. In our approach, the style information from the filtered source image is learned in adaptive manner, and normalized its feature maps in the encoder to be able to generate unfiltered images with the help of adversarial learning. Our main contribution in this study can be summarized as follows:
\begin{itemize}
    \item We propose a novel filter removal architecture, namely \textit{Instagram Filter Removal Network (IFRNet)}, which has encoder-decoder structure that normalizes the style information in the encoder to remove the visual effects of Instagram filters.
    \item We introduce a new dataset, namely \textit{Instagram-Filtered Fashionable Images (IFFI)}, which is a set of 10,200 high-resolution fashionable images composed of 600 collected images and their filtered versions by 16 different Instagram filters. 
    \item We compare the qualitative and quantitative results of IFRNet with the previous filter removal approach \cite{artisticFilter} and the fundamental image-to-image translation studies \cite{pix2pix2017, sidorov2019conditional, CycleGAN2017}. 
    \item As additional experiments, we present the filter classification performance of IFRNet, and analyze the dominant color estimation on the images unfiltered by the compared methods.
\end{itemize}


\section{Related Works}

\subsection{Filter Recognition and Removal} There are limited studies on recognizing the filters applied to an image. Chen \etal \cite{filterInvariant} introduces a Siamese CNN architecture to classify the filters by employing a discriminative pair sampling and an adaptive margin contrastive objective function. Bianco \etal \cite{bianco2017artistic} investigates the performance of some standard CNN architectures (\textit{i.e} AlexNet \cite{NIPS2012_c399862d}, LeNet \cite{Lecun98gradient-basedlearning} and GoogLeNet \cite{43022}) on small subset of Places2 dataset \cite{zhou2017places} where 22 Instagram filters are applied to the samples. Likewise, Chu and Fan \cite{chu2019photo} examines the performance of transfer learning approach on AlexNet \cite{NIPS2012_c399862d}, VGG16 \cite{Simonyan15} and ResNet-50 \cite{He2015} architectures. Sen \etal \cite{photoTransform} proposes a method using CNNs that learns to extract the parameters of transformations of filters from a reference image, and then transfers this information to a target image. Wu \etal \cite{Wu_Wu_Singh_Davis_2020} demonstrates that it is possible to reduce the effects of filtering on image classification by employing adaptive feature normalization approach. Lastly, Bianco \etal \cite{artisticFilter} discusses a noticeable strategy to remove the filters from the images. This strategy involves learning the parametric local transformations for each filter adaptively by CNNs to restore the images. Apart from these studies, we introduce an adversarial methodology that directly learns to remove the visual effects brought by the filters, and recover the images back to their original versions.

\subsection{Style Transfer}

Style Transfer is the variant of image-to-image translation tasks where the main goal is to transfer the style information extracted from a reference image into a target image while preserving its context information. Recent studies \cite{Gatys2015c, 46163, huang2017adain, pix2pix2017, Johnson2016Perceptual, attribute_hallucination, CycleGAN2017} show that \textit{many} style information from different reference images can be successfully transferred into \textit{many} different target images in varied domains. The underlying common strategy in these studies is to capture the style information from the feature representation of an image, and then to learn to synthesize this information and the context information of another image. Inspired from this strategy, we assume that the filters applied to images can be interpreted as the style information injected to the original version, and with the help of adaptive feature normalization \cite{huang2017adain}, it can be swept away from the images during the feature extraction. Therefore, we refer this task as \textit{reverse style transfer} where the particular style is removed from an image, instead of transferring it into a target image.

\captionsetup[subfigure]{labelfont=bf, labelformat=parens}
\begin{figure*}[!t]
        \centering
        \begin{subfigure}{0.138\textwidth}
                \includegraphics[width=\textwidth]{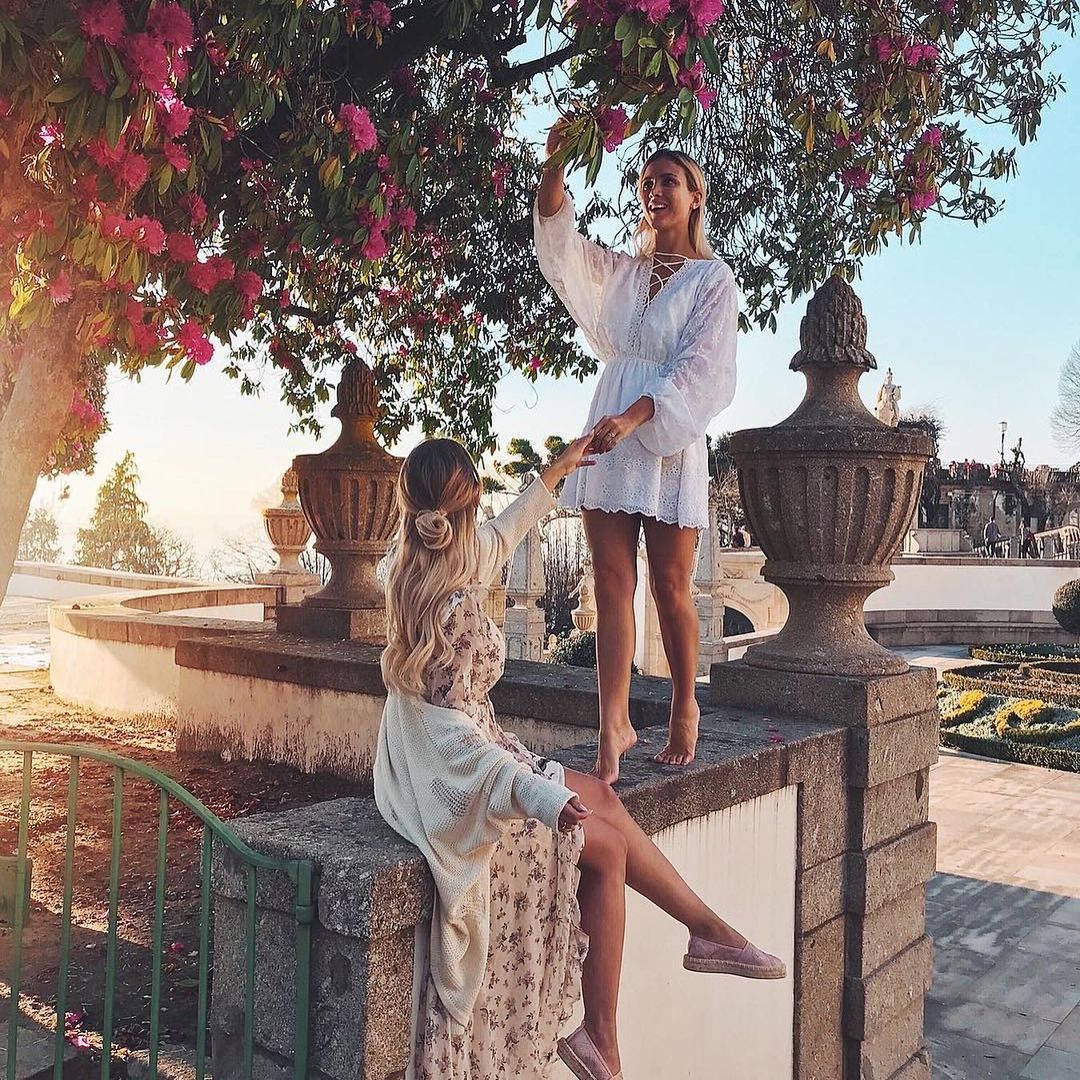}
                \includegraphics[width=\textwidth]{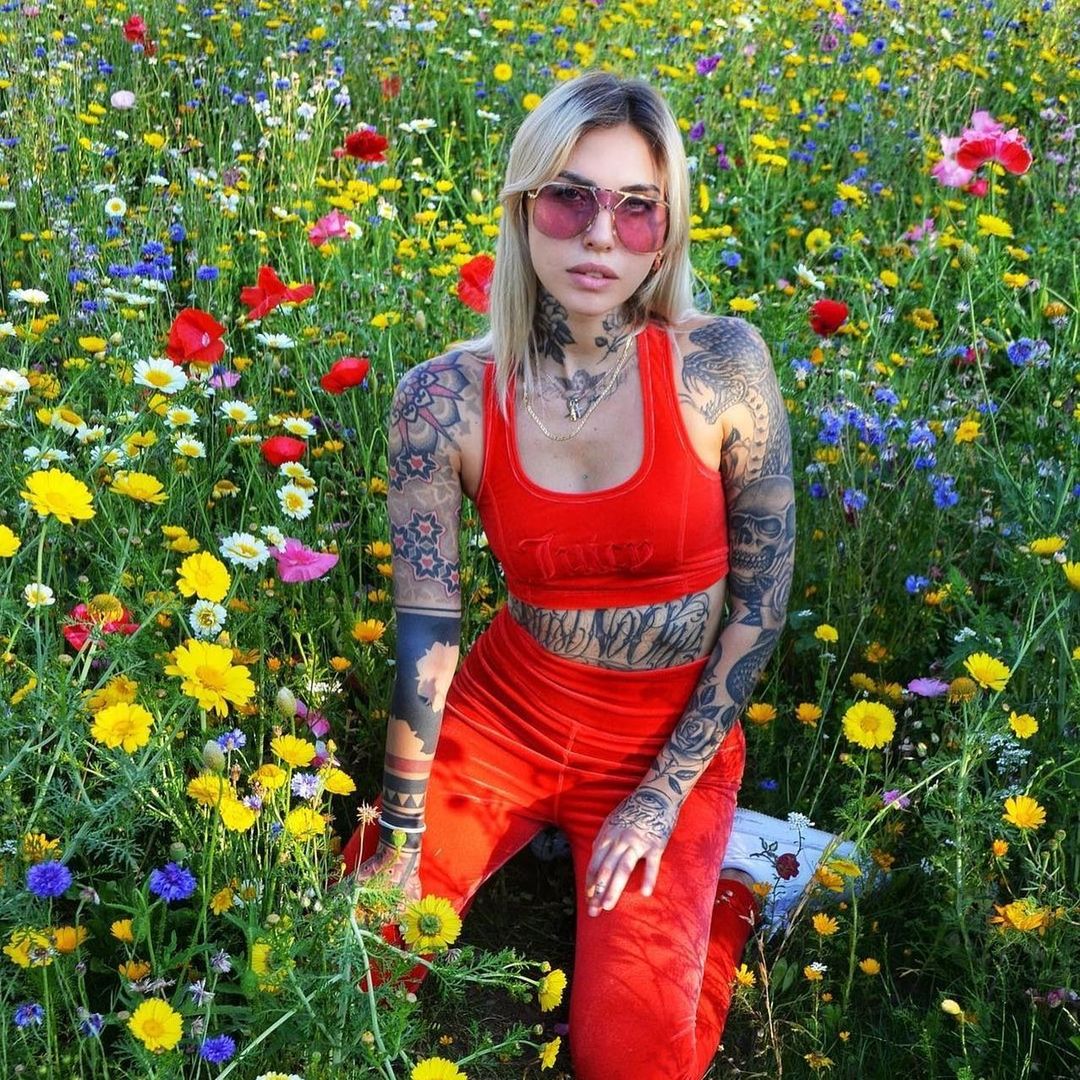}
                \includegraphics[width=\textwidth]{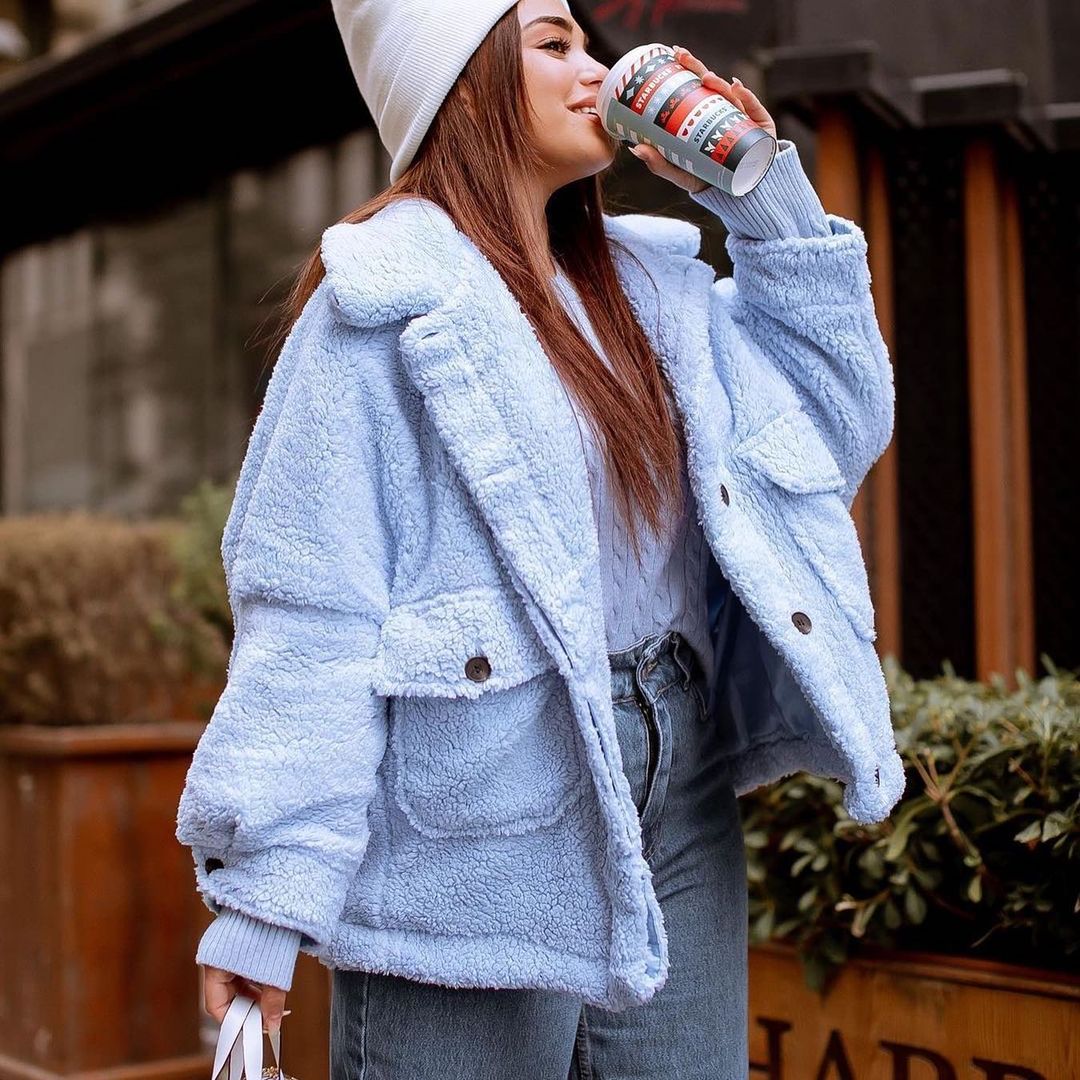}
                \caption{Original}
                \label{fig:fig2-org}
        \end{subfigure}       
        \begin{subfigure}{0.138\textwidth}
                \includegraphics[width=\textwidth]{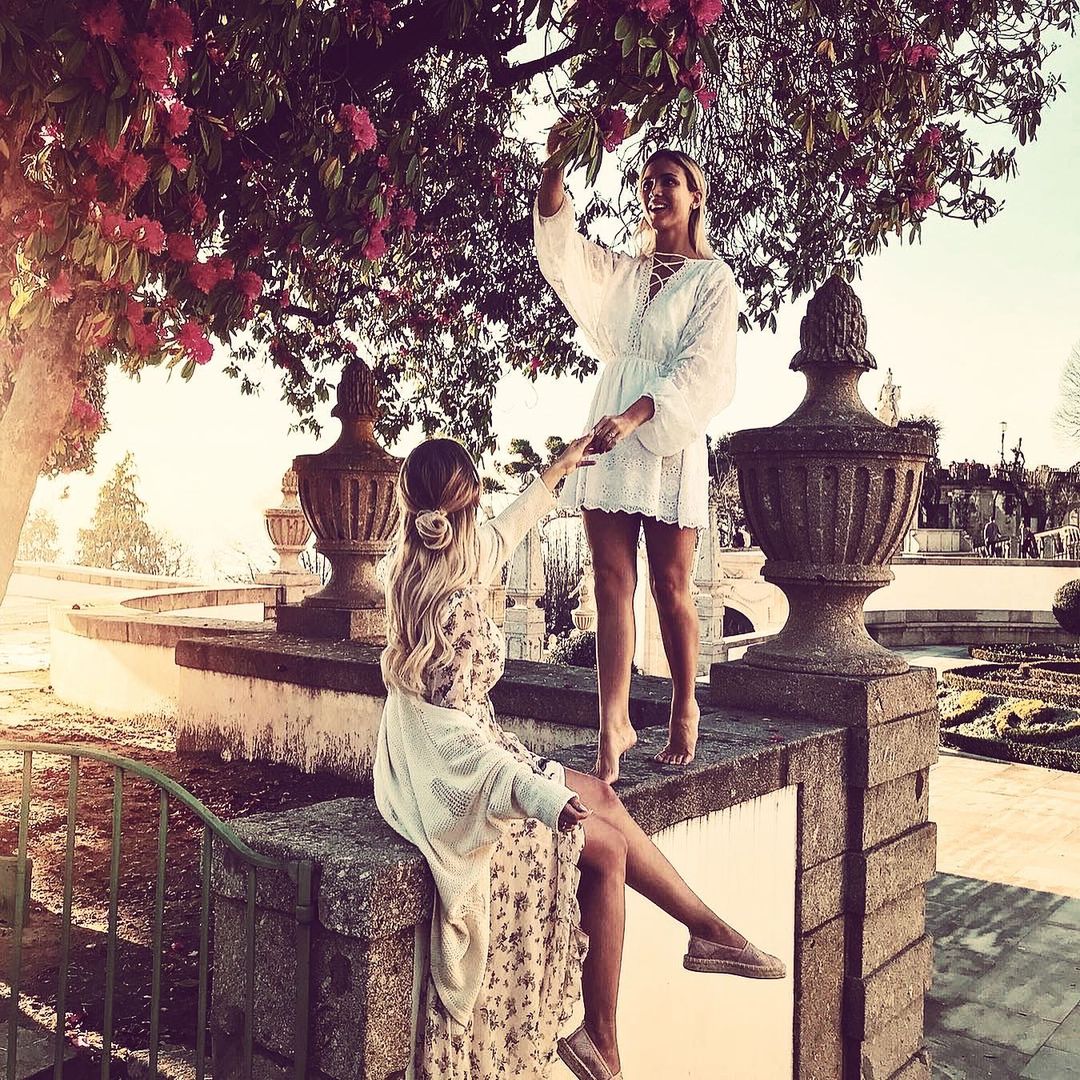}
                \includegraphics[width=\textwidth]{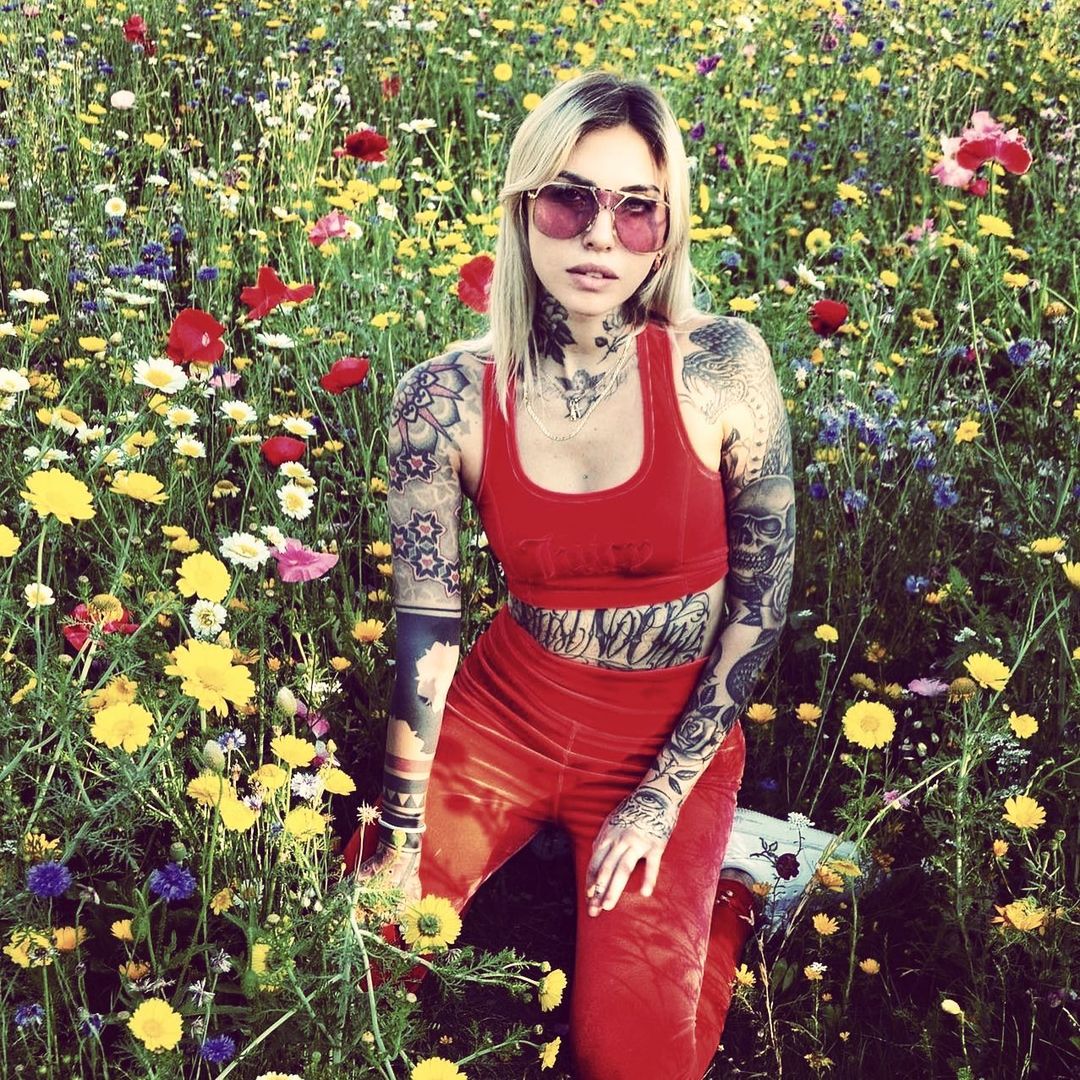}
                \includegraphics[width=\textwidth]{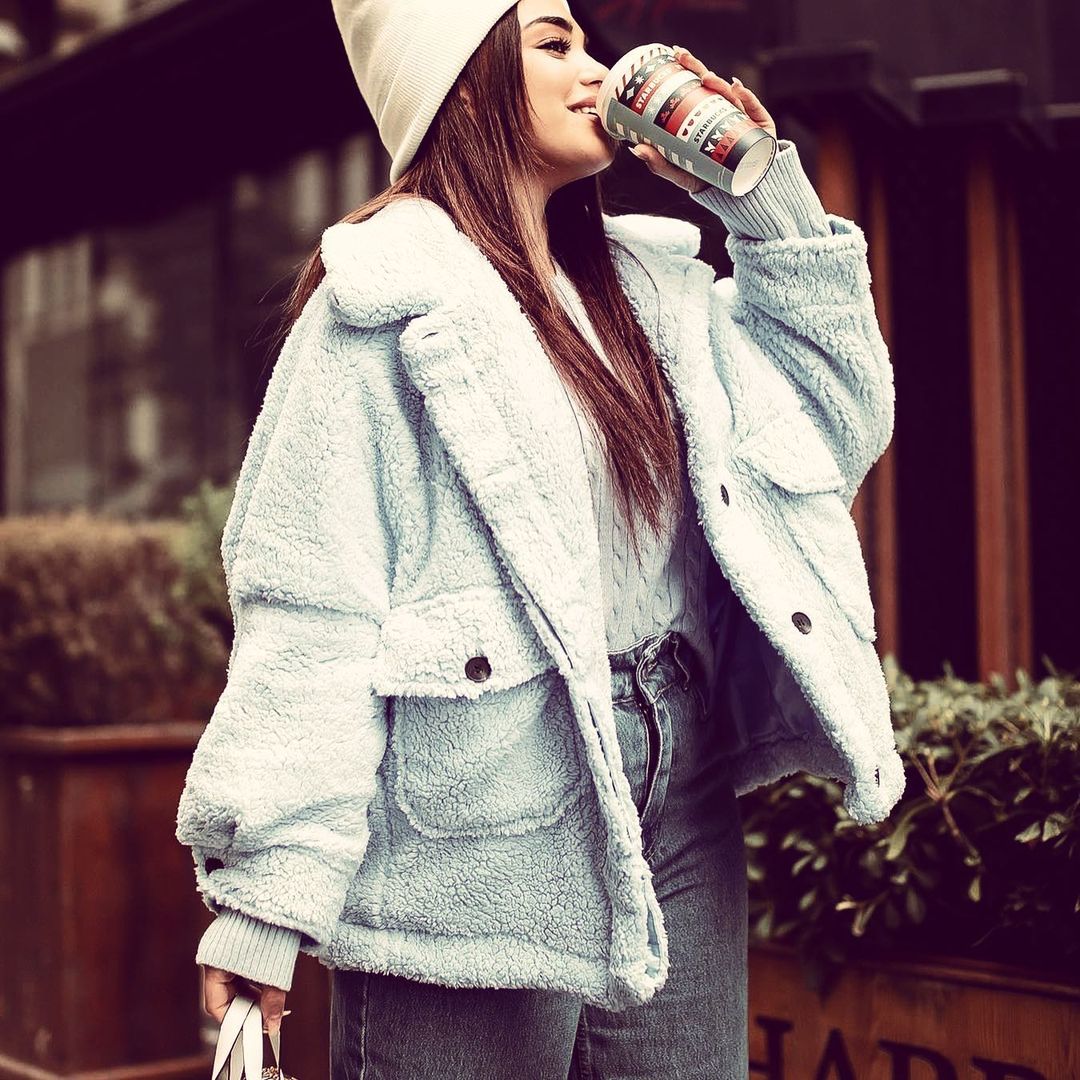}
                \caption{Brannan}
                \label{fig:fig2-brannan}
        \end{subfigure}
        \begin{subfigure}{0.138\textwidth}
                \includegraphics[width=\textwidth]{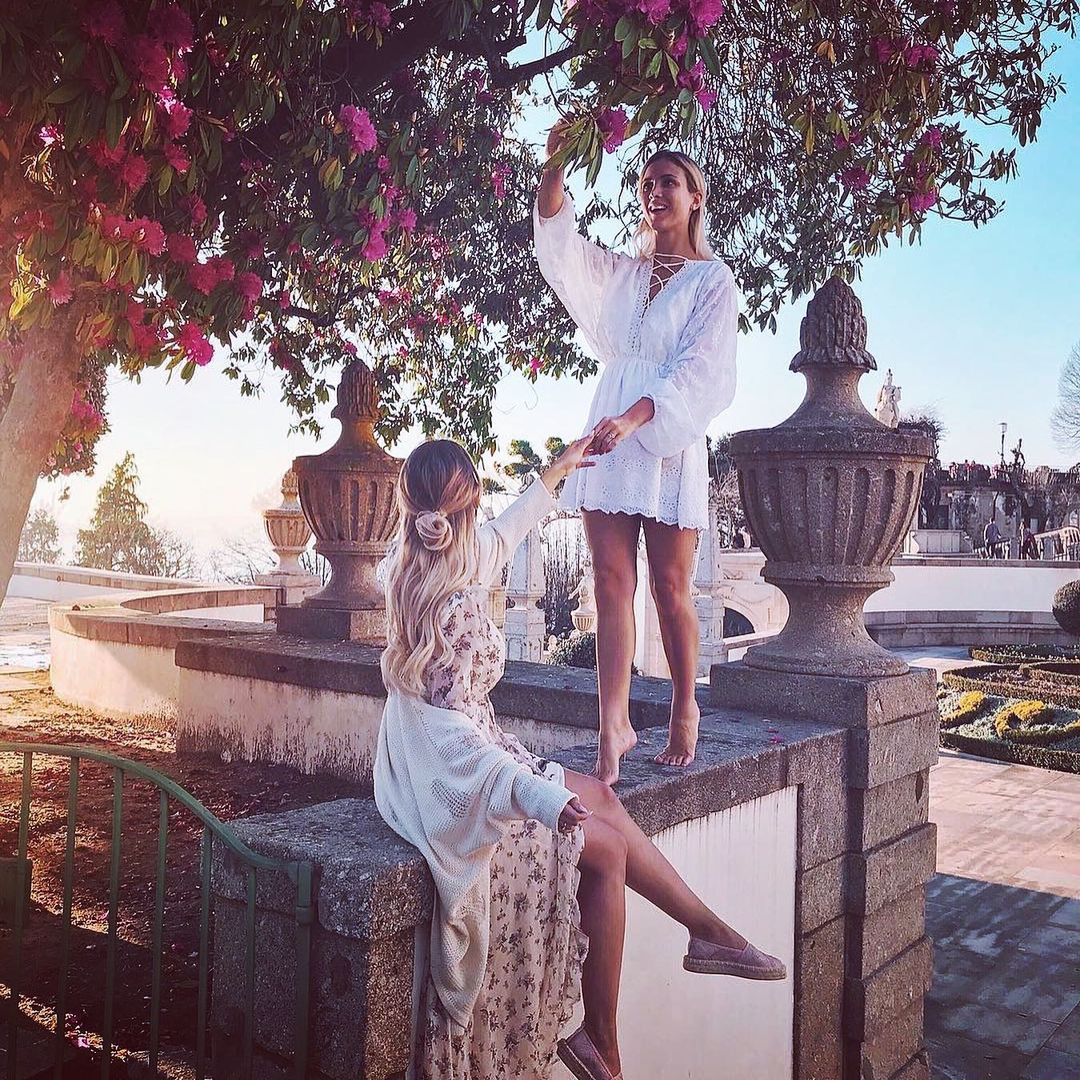}
                \includegraphics[width=\textwidth]{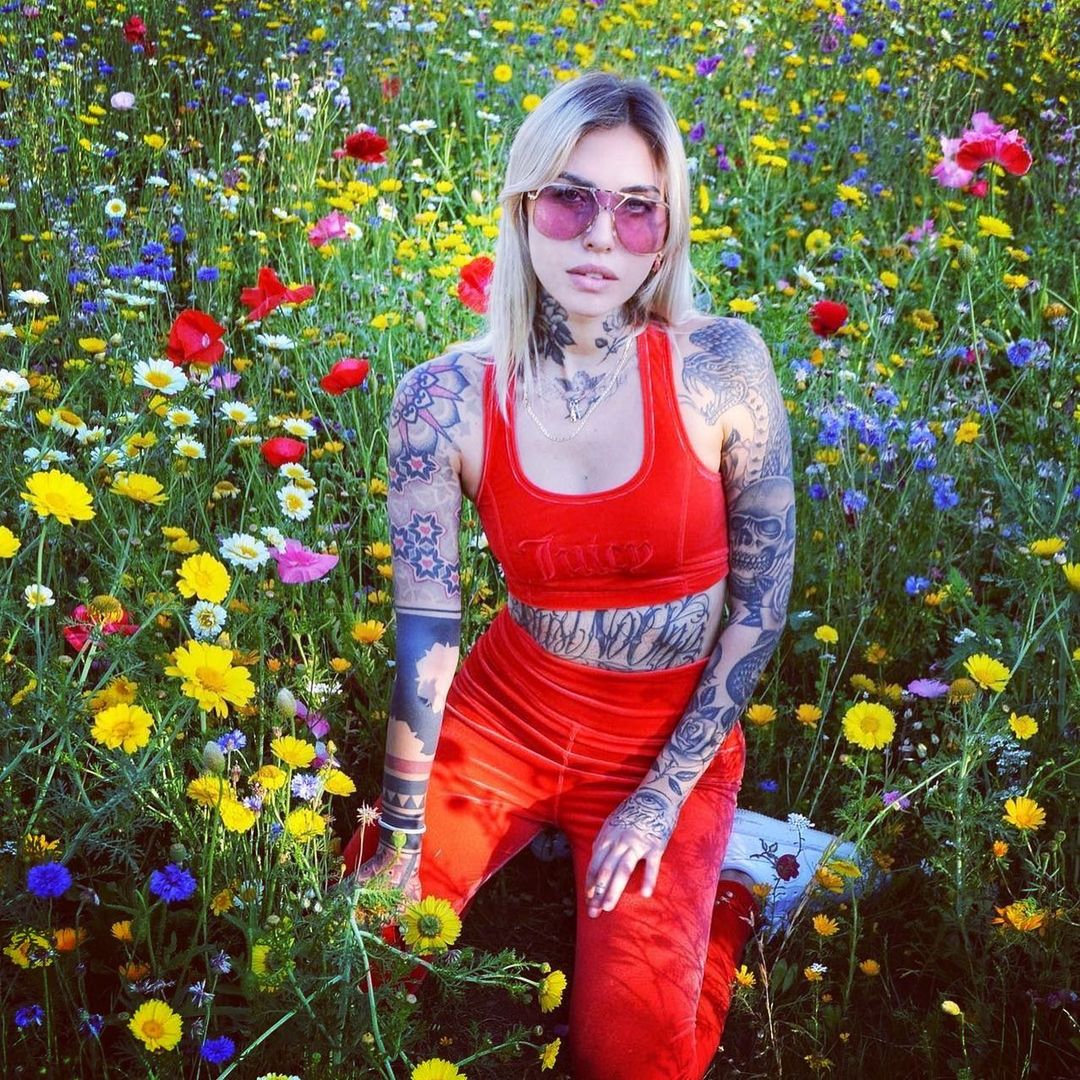}
                \includegraphics[width=\textwidth]{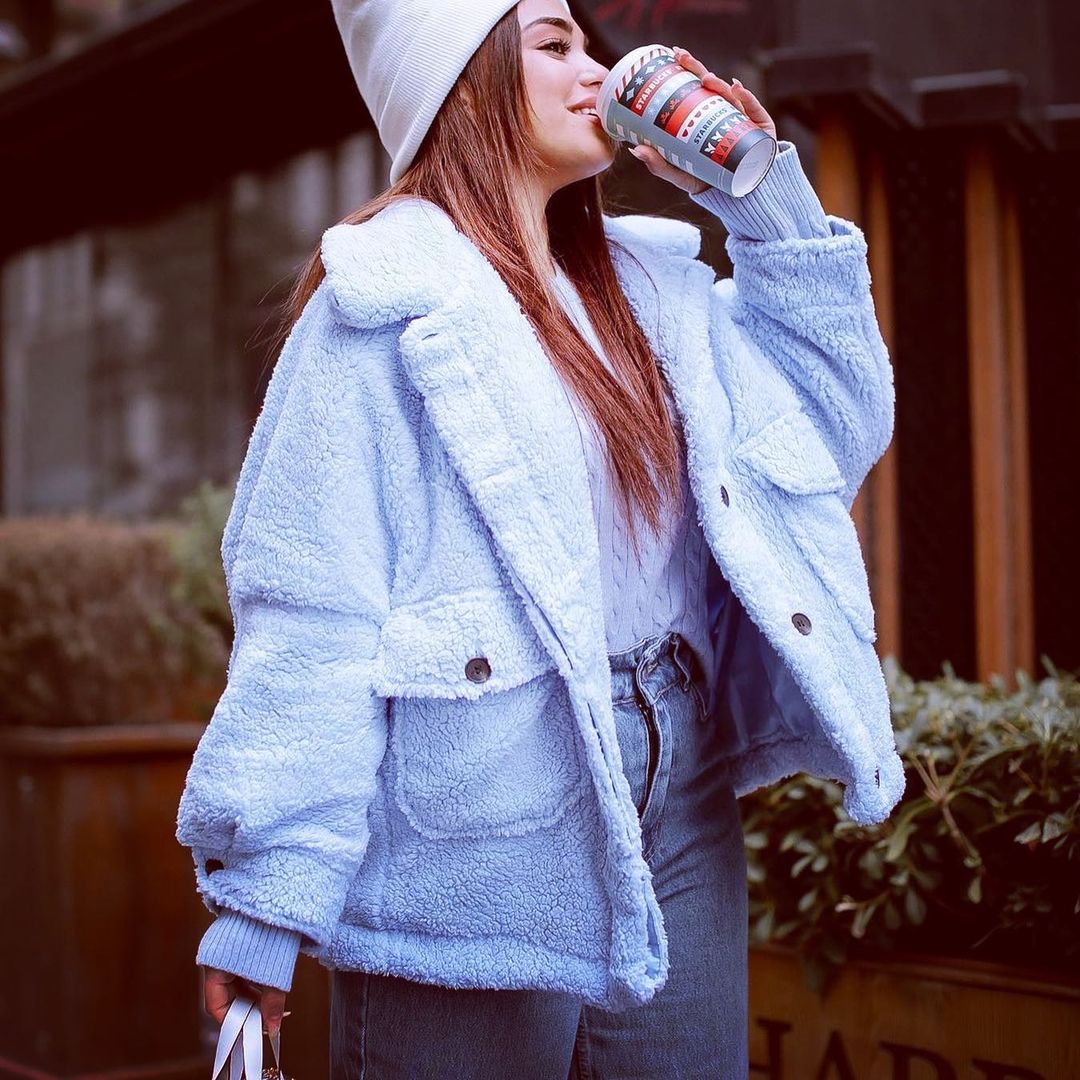}
                \caption{Hudson}
                \label{fig:fig2-hudson}
        \end{subfigure}
        \begin{subfigure}{0.138\textwidth}
                \includegraphics[width=\textwidth]{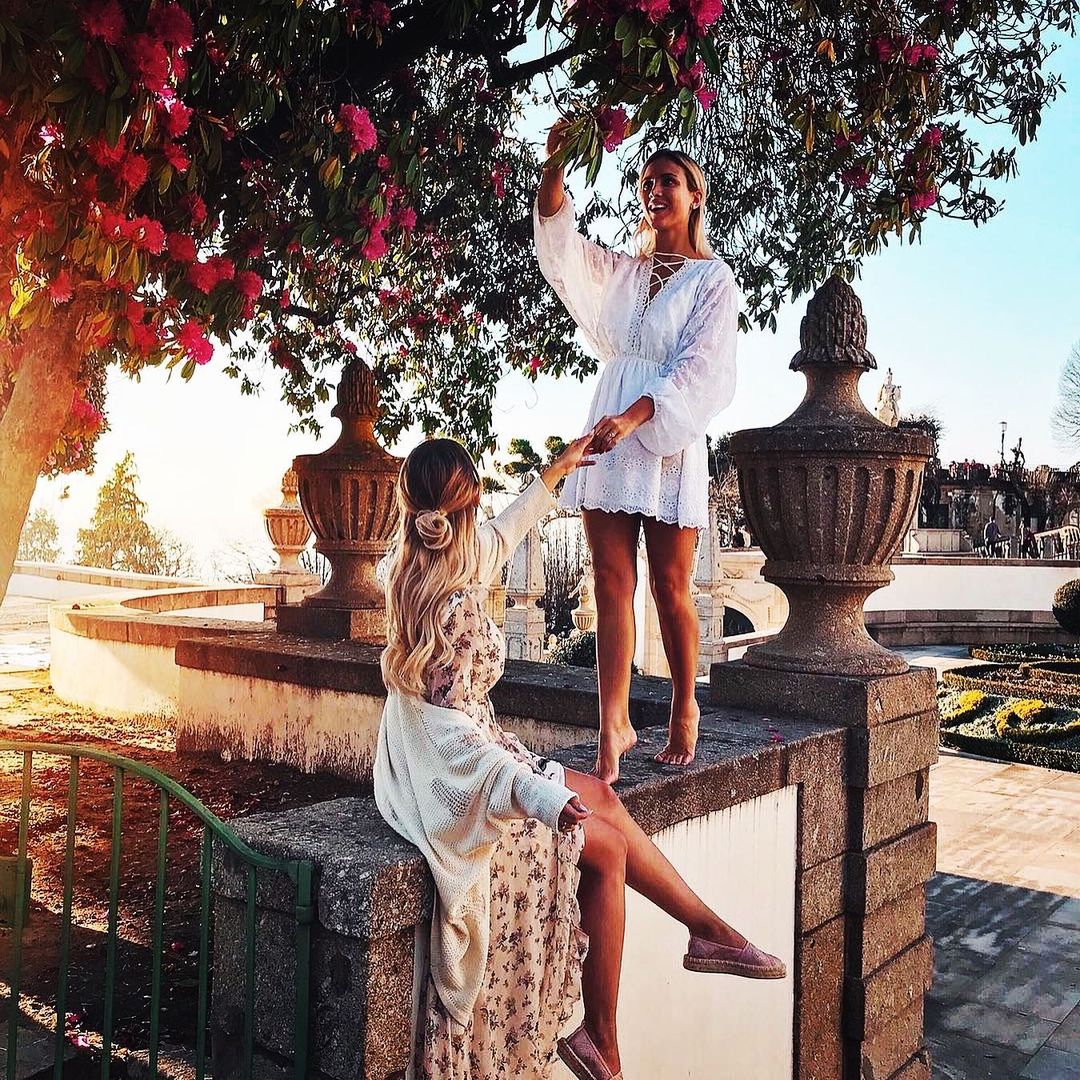}
                \includegraphics[width=\textwidth]{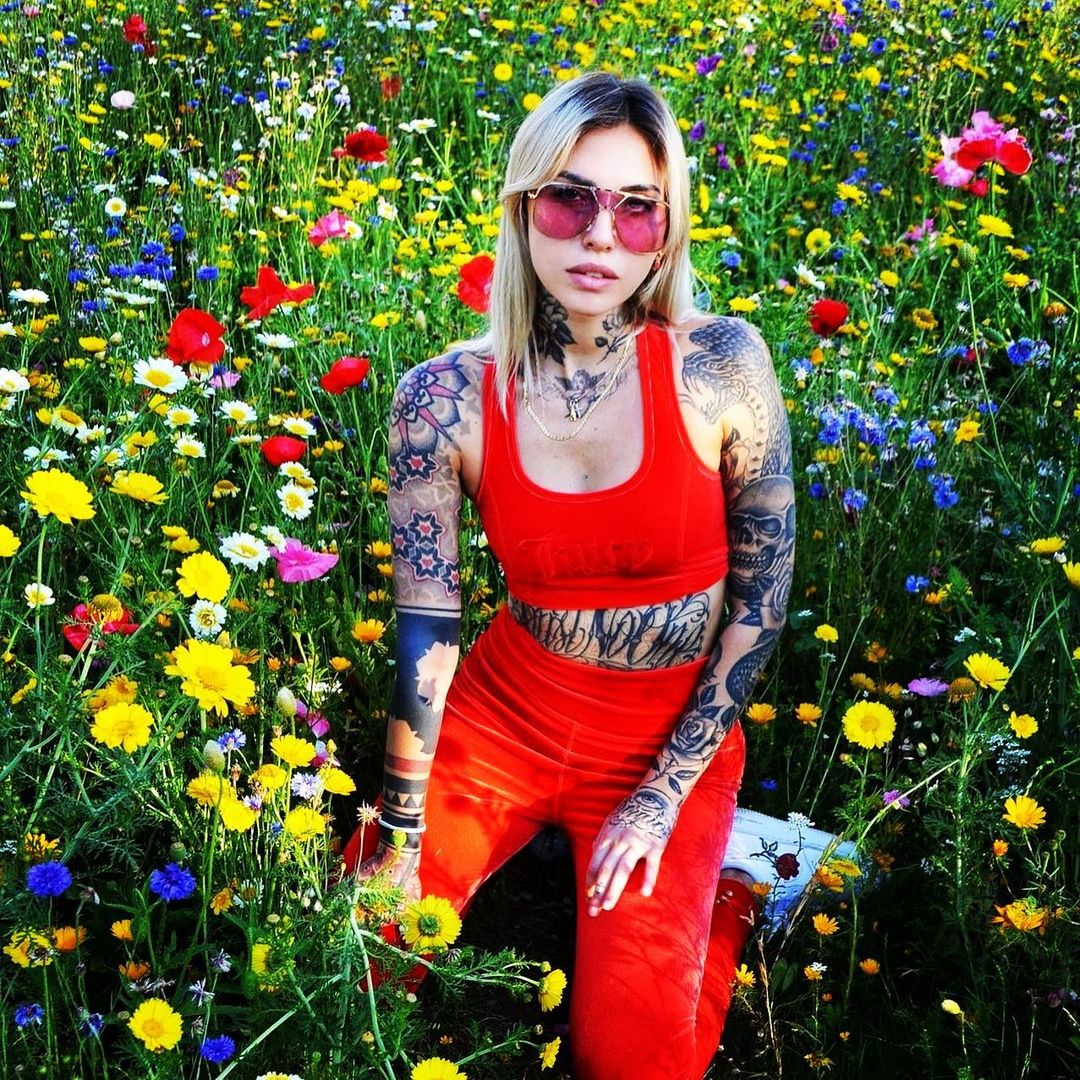}
                \includegraphics[width=\textwidth]{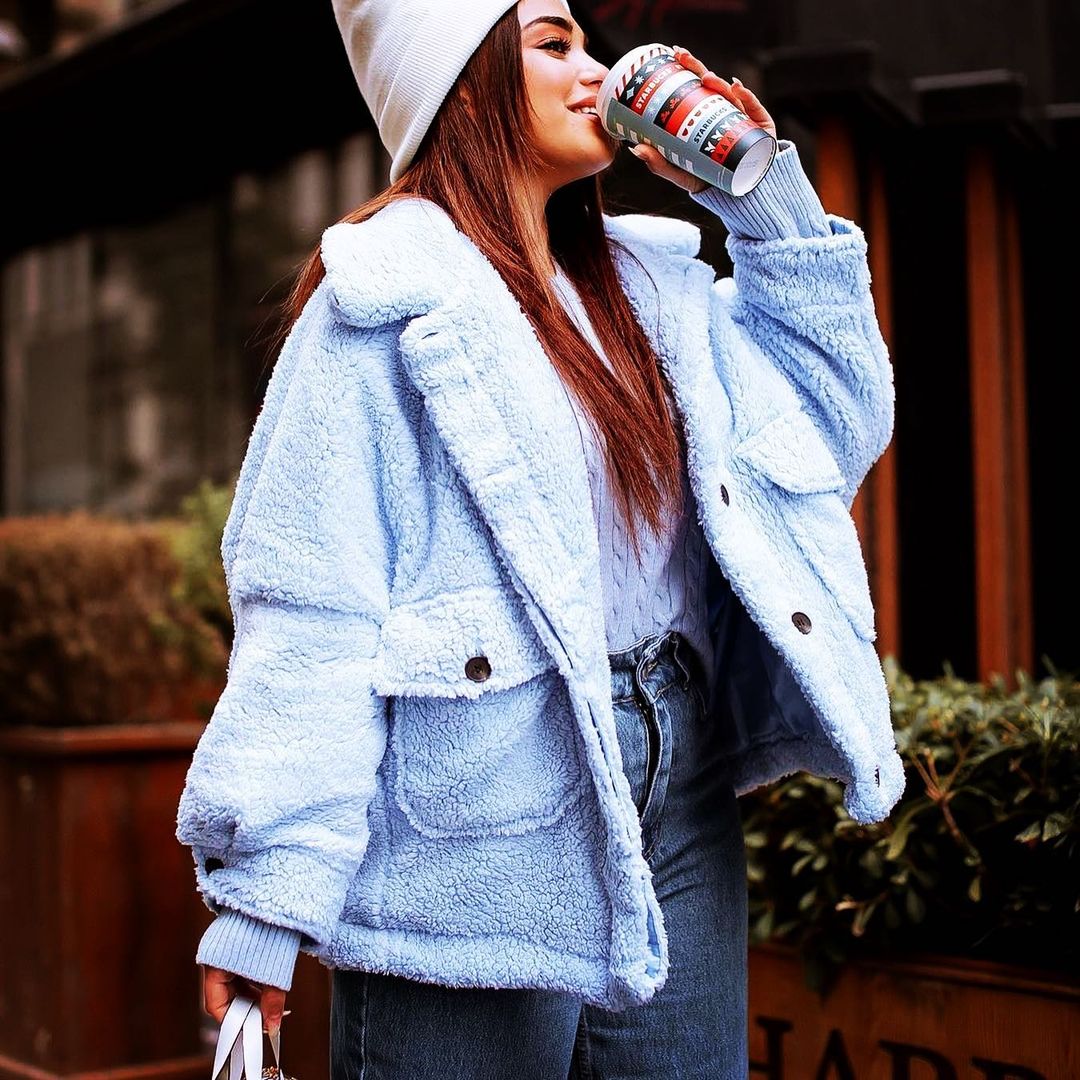}
                \caption{Lo-Fi}
                \label{fig:fig2-lofi}
        \end{subfigure}
        \begin{subfigure}{0.138\textwidth}
                \includegraphics[width=\textwidth]{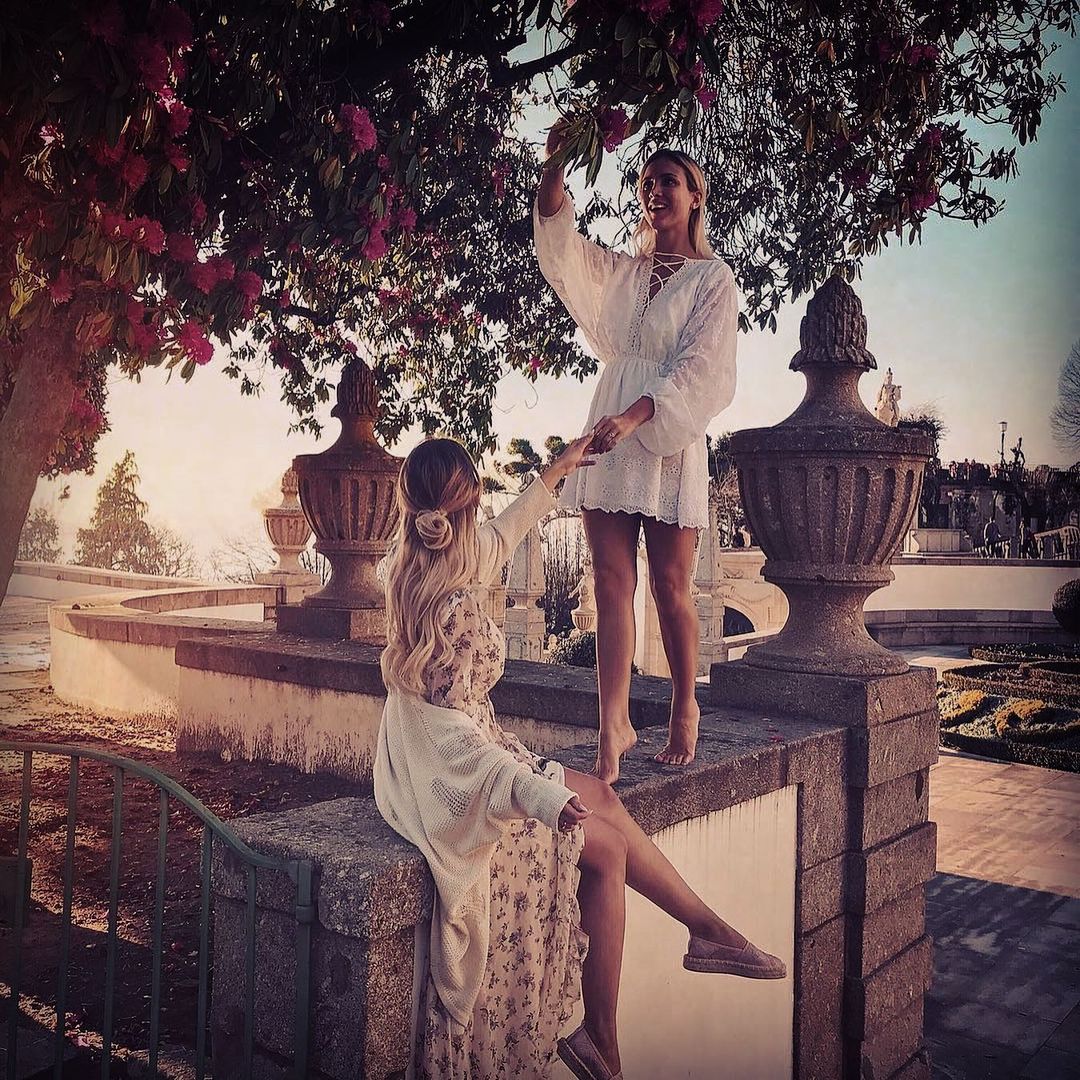}
                \includegraphics[width=\textwidth]{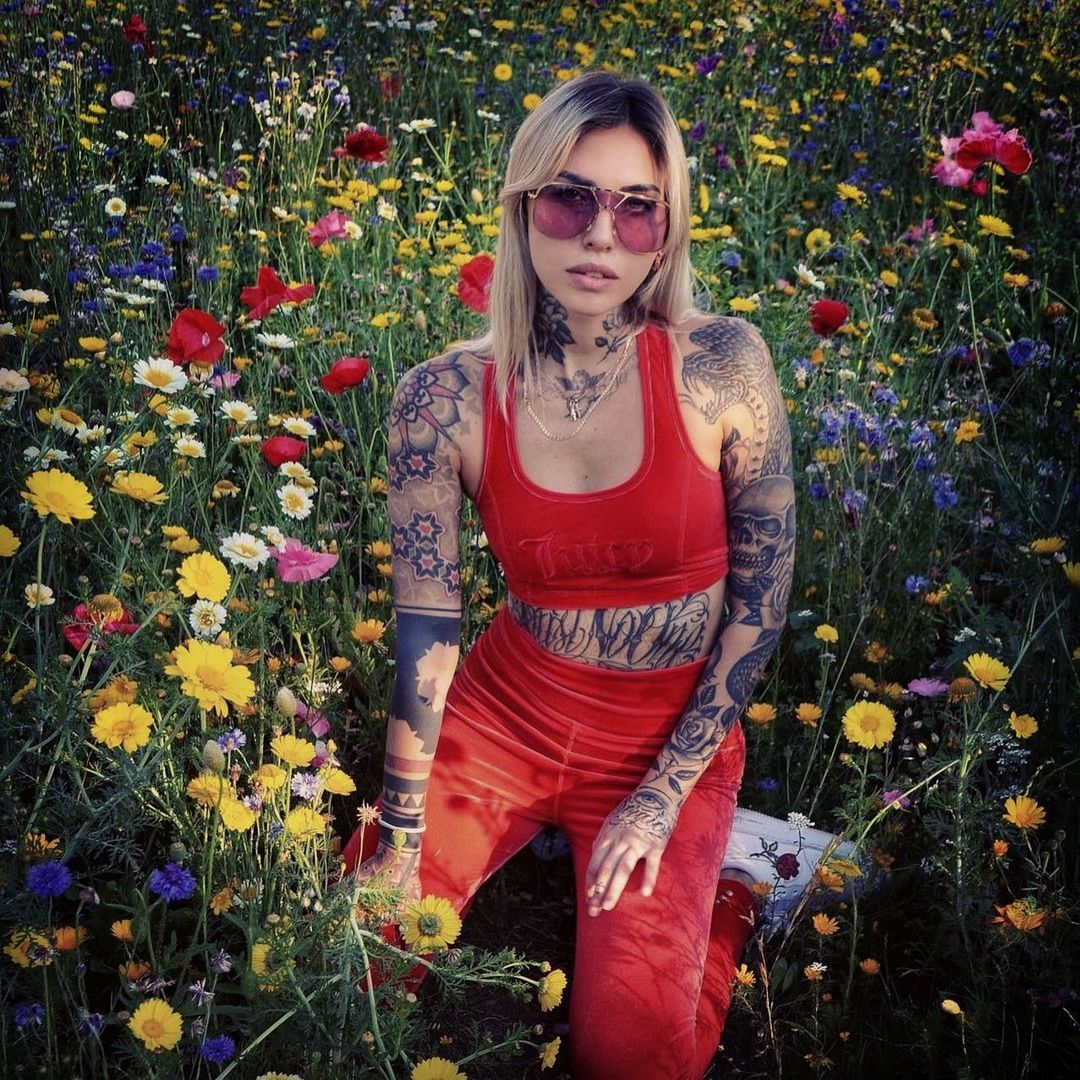}
                \includegraphics[width=\textwidth]{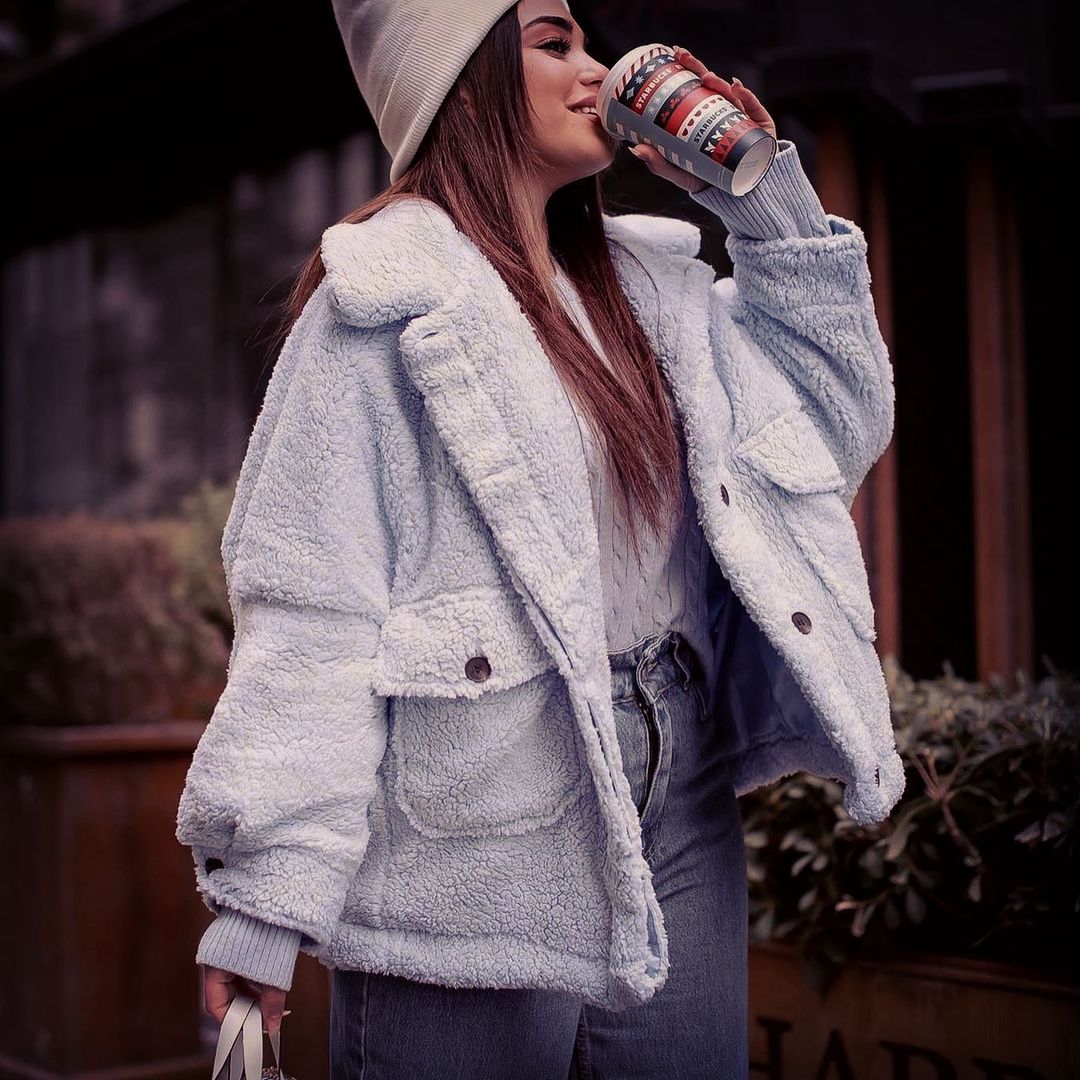}
                \caption{Sutro}
                \label{fig:fig2-sutro}
        \end{subfigure}
        \begin{subfigure}{0.138\textwidth}
                \includegraphics[width=\textwidth]{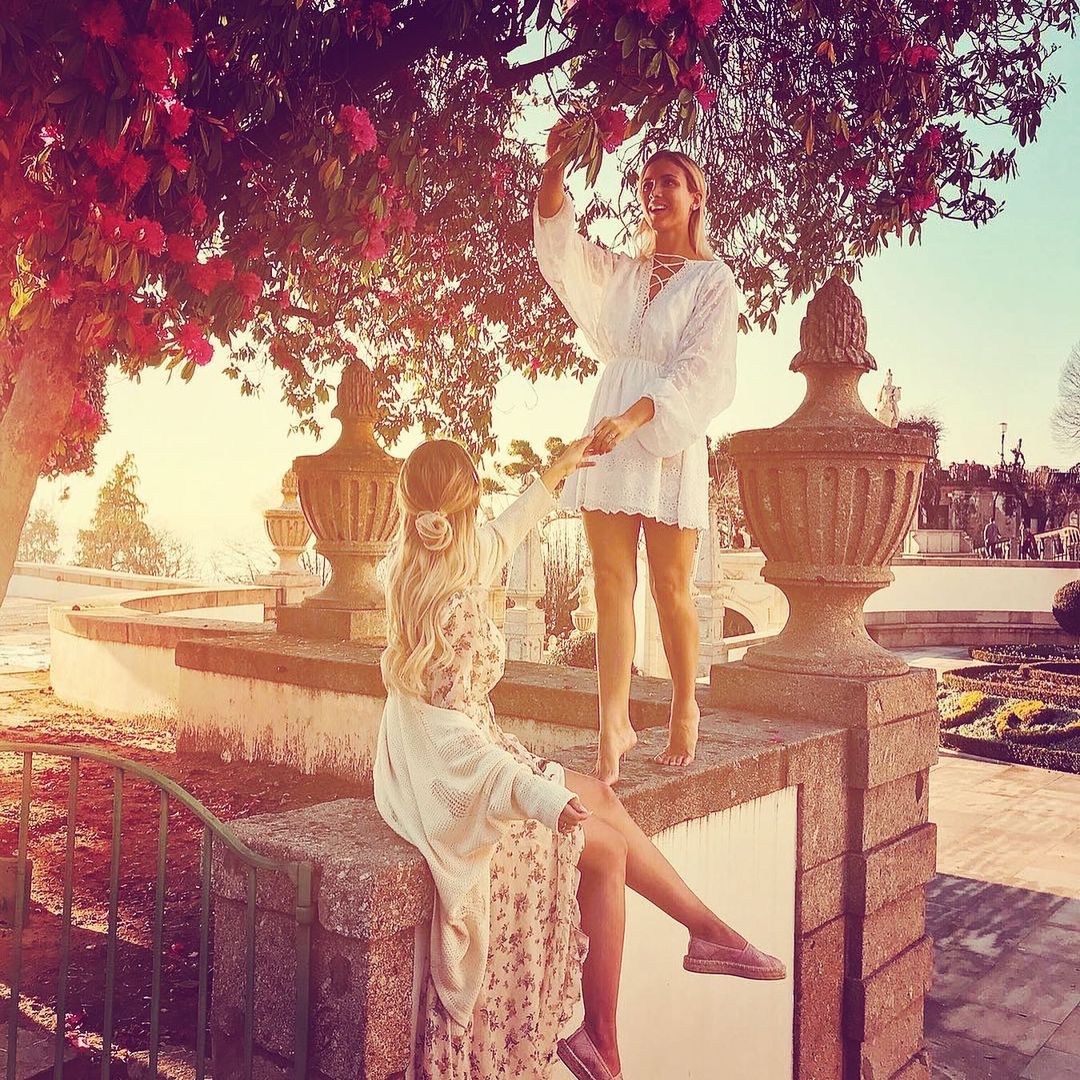}
                \includegraphics[width=\textwidth]{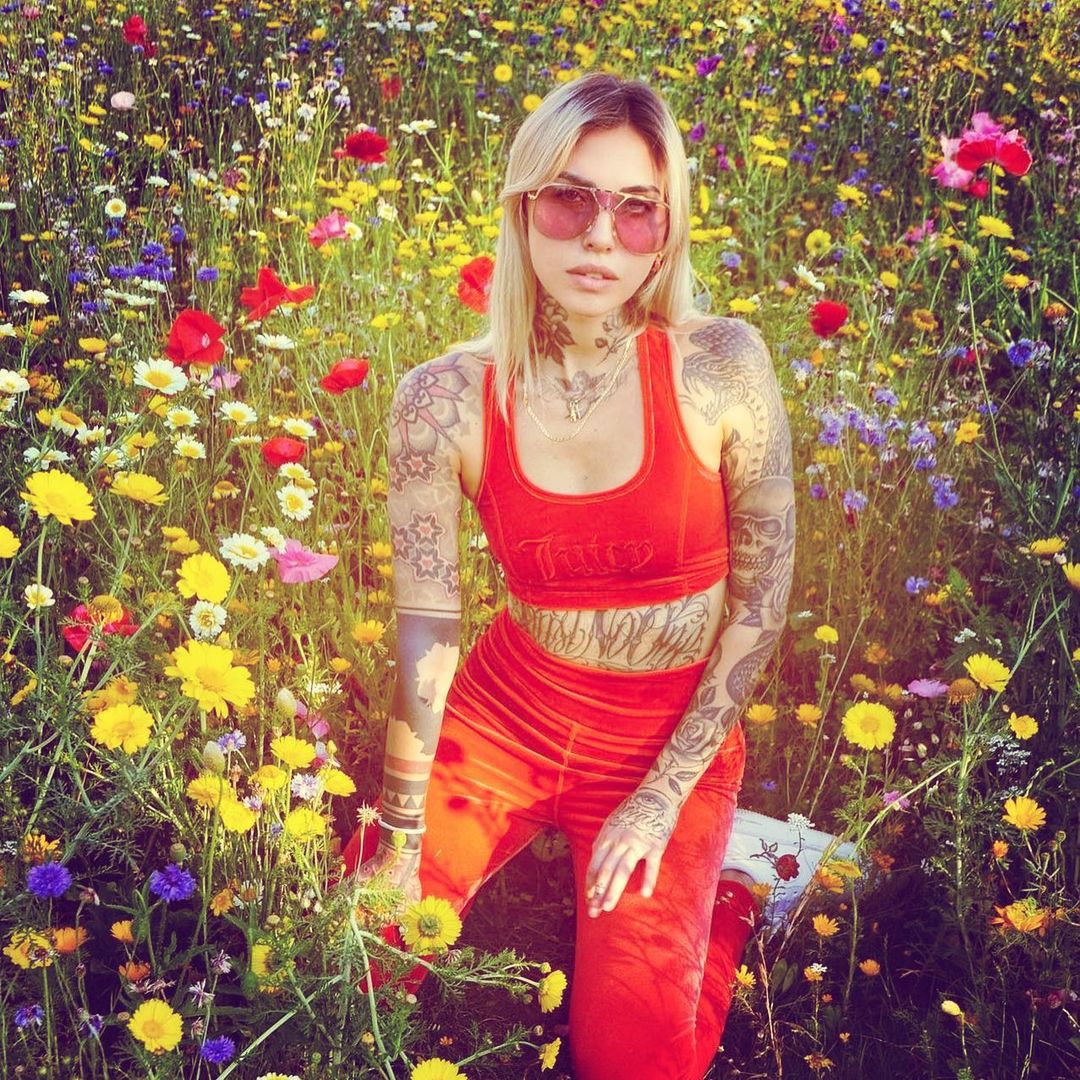}
                \includegraphics[width=\textwidth]{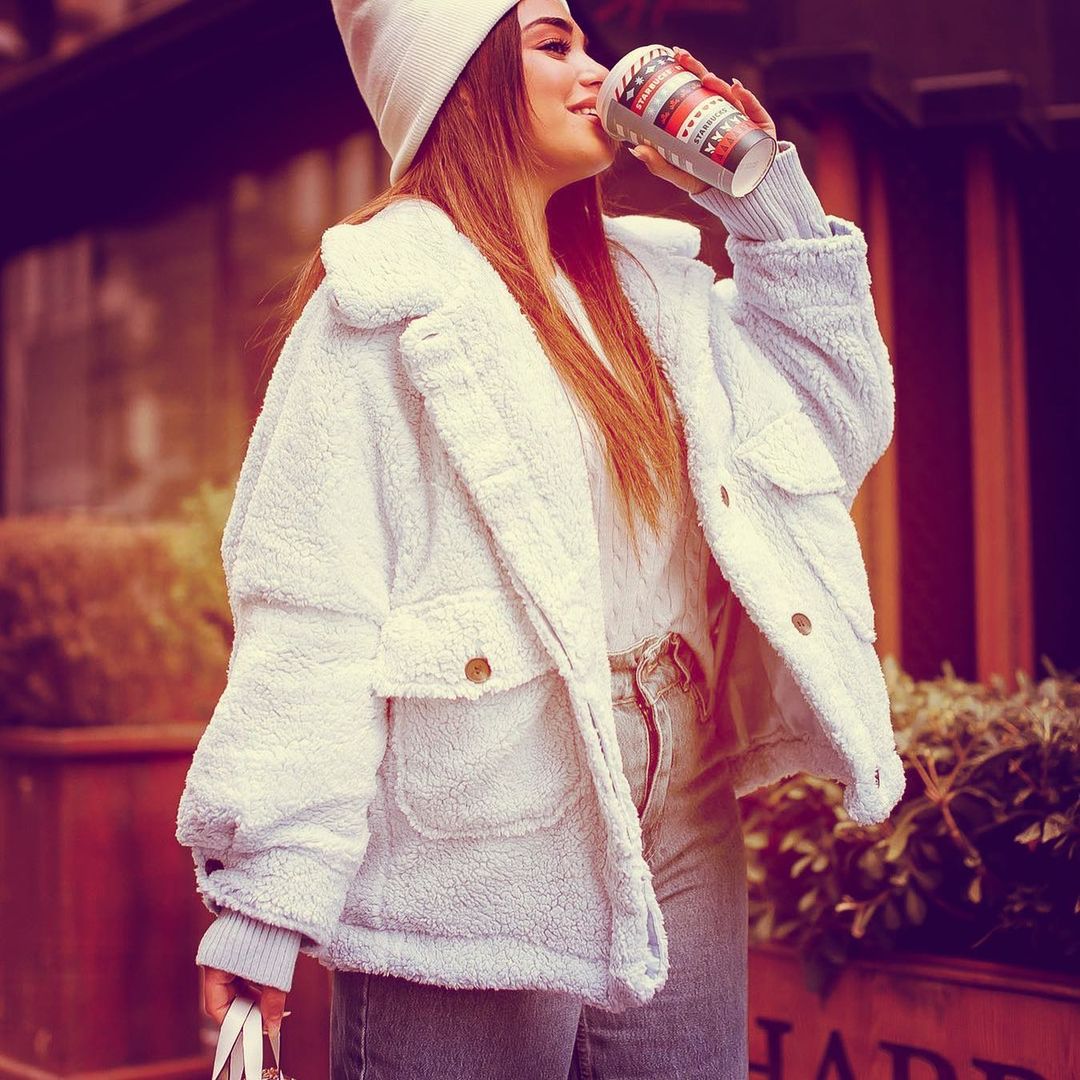}
                \caption{Toaster}
                \label{fig:fig2-toaster}
        \end{subfigure}
        \begin{subfigure}{0.138\textwidth}
                \includegraphics[width=\textwidth]{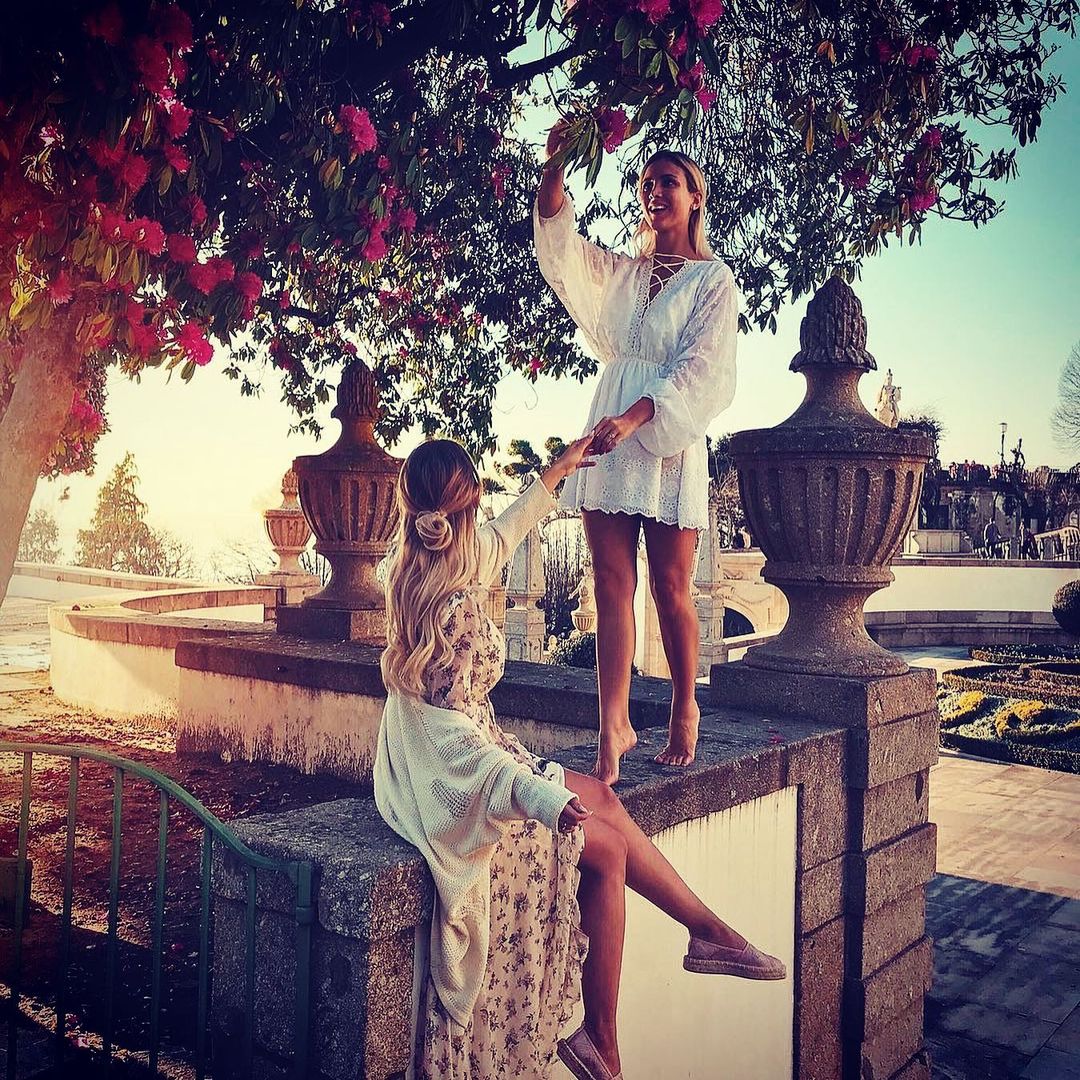}
                \includegraphics[width=\textwidth]{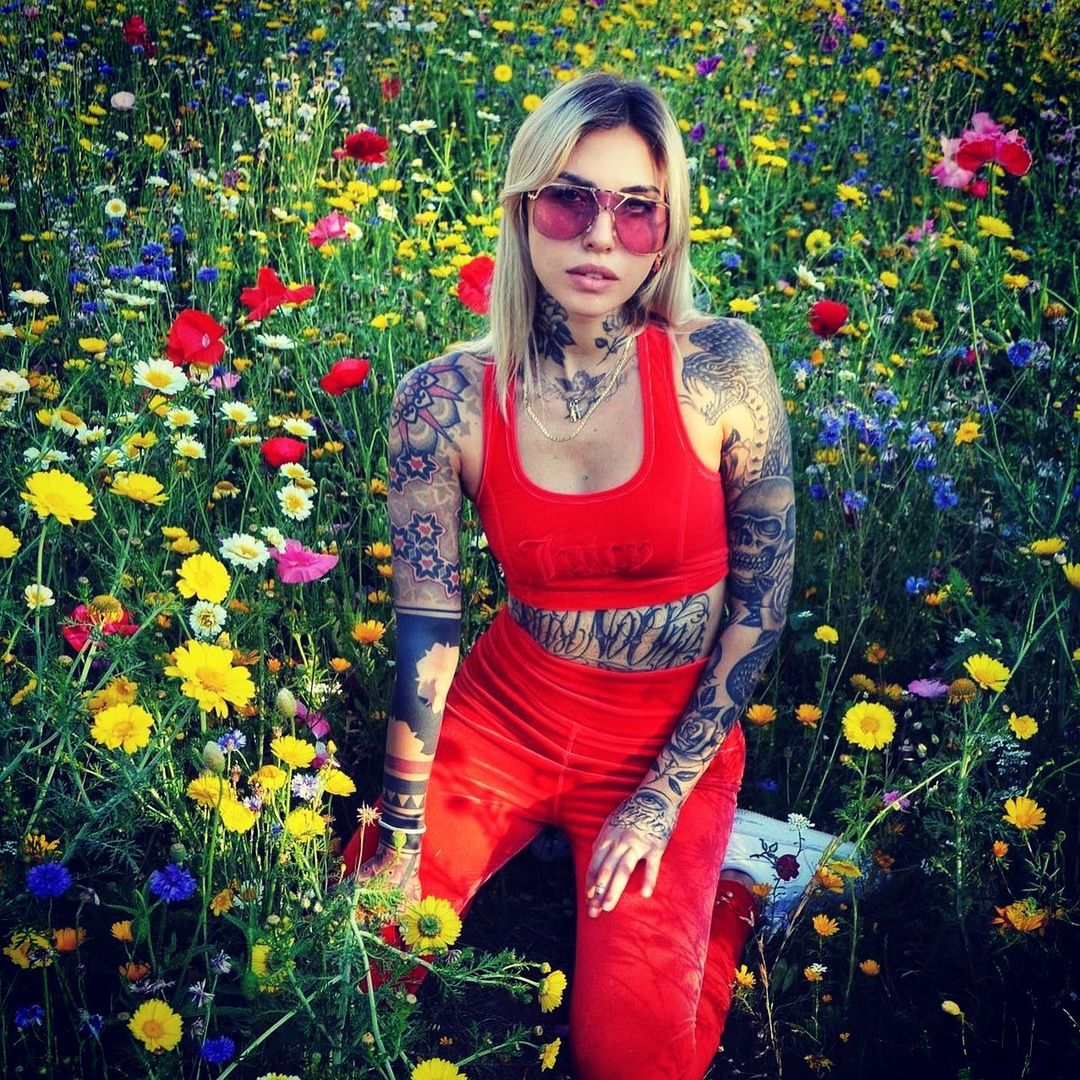}
                \includegraphics[width=\textwidth]{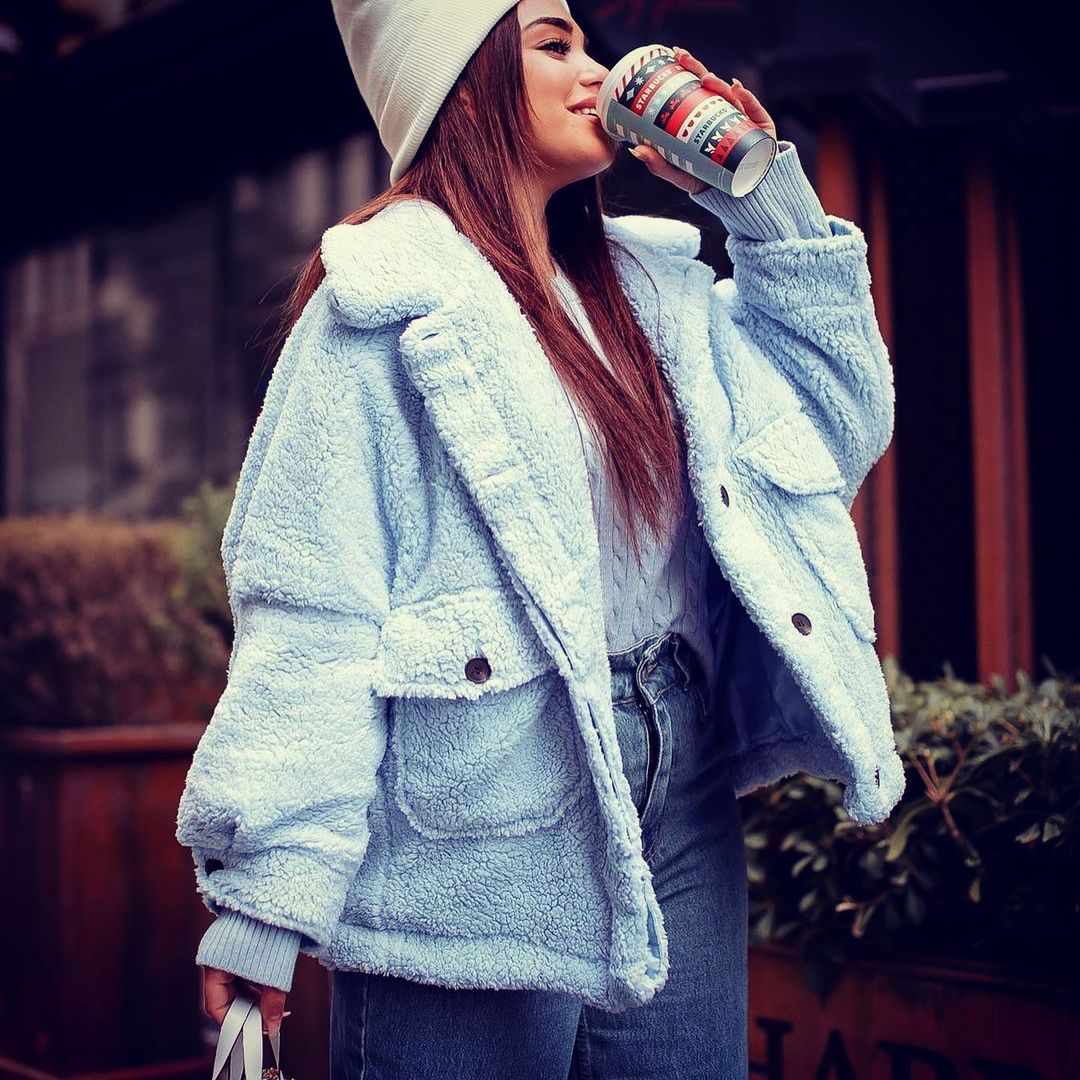}
                \caption{X-Pro II}
                \label{fig:fig2-xpro}
        \end{subfigure}
        
        \caption{Original images from IFFI dataset and their filtered versions by 7 example filters.}\label{fig:fig2} 
\end{figure*}

\section{Methodology}

Our proposed architecture, namely \textit{IFRNet}, directly removes the style information injected to the images by filters. IFRNet achieves it by normalizing the feature maps of filtered images, and directly generates the unfiltered version of them. Moreover, we introduce a dataset containing 10,200 filtered high-resolution fashionable images to validate the performance of our approach.

\subsection{IFFI: Instagram-Filtered Fashionable Images}

Previous studies have picked a small subset of well-known datasets (\textit{i.e.} \cite{imagenet_cvpr09, zhou2017places}) to use for validating their performance. These subsets with the filtered versions of the samples have not been open-sourced, and it may not be possible to reproduce the results in these studies. Moreover, the samples in these datasets do not seem like social media posts, which mostly have aesthetic concerns. Due to these reasons, we have collected a set of aesthetically pleasing images that are filtered by 16 Instagram filters. \textit{IFFI} dataset contains high-resolution ($1080 \times 1080$) 600 images and with their 16 different filtered versions for each. In particular, we have picked mostly-used 16 filters: \textit{1977, Amaro, Brannan, Clarendon, Gingham, He-Fe, Hudson, Lo-Fi, Mayfair, Nashville, Perpetua, Sutro, Toaster, Valencia, Willow, X-Pro II}. Some examples of collected images and their filtered versions can be seen in Figure \ref{fig:fig2}.

\subsection{IFRNet: Instagram Filter Removal Network}
\label{sec:ifr}

We define the problem of removing Instagram filters from the images as a reverse style transfer problem, where any visual effect injected by a particular filter is removed from an image by directly reverting them back to its original style. To achieve this, we propose \textit{Instagram Filter Removal Network (IFRNet)}, which has encoder-decoder structure employing adaptive feature normalization strategy to all layers in the encoder. 

Style extractor module contains five-layer fully-connected network $f_{fc}$. It maps the feature representations $\mathbf{z}_{vgg}$, encoded by a pre-trained VGG network, to the latent space. There are $N$ fully-connected heads attached to $f_{fc}$, where $N$ equals to the number of normalization layer in the encoder. Each head is responsible for adapting the affine parameters $y_{i}$ of corresponding normalization layers in the encoder. This module can be formulated as follows

\begin{equation}
    y_{i} = h_{i}(f_{fc}(\mathbf{z}_{vgg})) 
\end{equation}
where $h_{i}(\cdot)$ is $i^{th}$ fully-connected head of style extractor module and $y_{i}$ represents the predicted mean and variance vectors for the corresponding normalization layer. 

Adaptive instance normalization (AdaIN) \cite{huang2017adain} makes it possible to adapt the style of an input image to any arbitrary style of a target image by transferring the feature statistics computed across spatial locations. Formally, AdaIN receives the feature maps of the content image $x$ and the style input $y$, and aligns the channel-wise mean and variance of these maps to match the style information obtained from $y$.

\begin{equation}
    \text{AdaIN}(x, y) = \sigma(y) \left(\frac{x - \mu(x)}{\sigma(x)}\right) + \mu(y)
\end{equation}

\begin{figure*}[!t]
    \centering
    \includegraphics[width=0.94\textwidth]{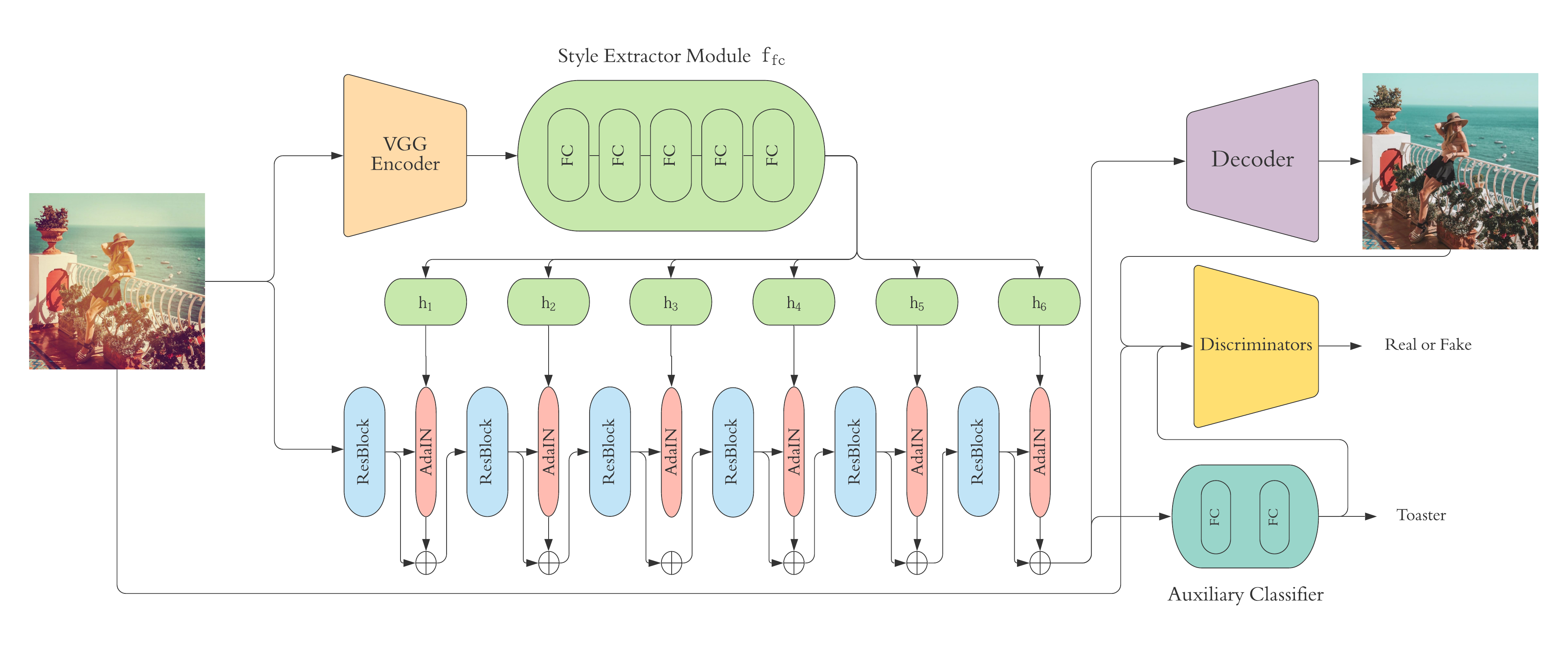}
    \caption{Overall architecture of IFRNet. The latent representations of the images are fed into style extractor module $f_{fc}$, then the affine parameters are calculated, and sent to the corresponding AdaIN layer to eliminate the style information from the feature maps. At the end, the decoder part of IFRNet generates the unfiltered version of the input images.}\label{fig:fig3} 
\end{figure*}

In our design, we consider the filters applied to the images as an external style information, and simply employ AdaIN to eliminate this information from these images. The encoder of IFRNet is composed of 6 residual blocks, and each of them has specific AdaIN layer to normalize the feature maps in each level with its corresponding affine parameters extracted by previous module. To preserve any related information about the original style, we include skip connections to the normalized feature maps before sending them to the next layer. In this way, the encoder has the capacity not only to learn to remove the external style information, but also to preserve the related ones.

\begin{equation}
    o_{i} = r_i(v_i, y_i) + v_i
\end{equation}
where $r_i(\cdot)$ represents the residual block at $i^{th}$ layer of the encoder, which receives the feature maps $v_i$ and the affine parameters $y_i$, and $o_i$ is the output of this block. Note that the normalization does not have any impact on the feature maps when its corresponding affine parameters are set to zero for a particular layer, which means there is no external style information remaining on feature maps at this level. 

After obtaining the latent representations of the filtered images including no external style information, we feed them to both the decoder part of IFRNet and an auxiliary classifier for filter type. The decoder contains a number of consecutive upsampling and residual convolutional blocks, and it generates the unfiltered version of the input image with the help of adversarial training. Moreover, the auxiliary classifier is a simple feed-forward network that produces the predicted filter type.  At this point, we design two discriminator models (\ie PatchGAN) which separately penalize the global image and the local structures at the scale of patches, and follow the adversarial training strategy in \cite{pix2pix2017}. Overall architecture of IFRNet is shown in Figure \ref{fig:fig3}. 

The objective function for IFRNet is composed of three main components, which are texture consistency loss, semantic consistency loss, adversarial loss extended with auxiliary classification loss. Texture consistency loss computes the sum of the patch-wise relative difference between the feature maps of the input and the target (\ie ID-MRF loss) \cite{mechrez2018Learning, mechrez2018contextual}, and leads to enhance the details in the generated images by minimizing the inconsistency between the input patch and the most similar target patch. Secondly, semantic consistency loss \cite{Johnson2016Perceptual} ensures the input and target images have similar feature representations at different scales by minimizing Euclidean distance between them (see Eq. \ref{eq:semantic}).

\begin{equation}
    \label{eq:semantic}
    \mathcal{L}_{sem} =\sum\limits_{p=0}^{P-1} \frac{1}{C_i H_i W_i}\| {\Phi_i^p(\mathbf{I}_{out}}) - {\Phi_i^p(\mathbf{I}_{gt}})\|_{2}^{2} 
\end{equation}
where $\Phi^p(\cdot)$ is the feature map of $p^{th}$ layer of pre-trained VGG16 network \cite{Simonyan15} with the shape of $C_i \times H_i \times W_i$, $\mathbf{I}_{out}$ is the output of IFRNet and $\mathbf{I}_{gt}$ is the input image. At this point, we use only \texttt{relu3\_2} layer of VGG16 for this loss with the intent of controlling over the computational complexity of training. Morover, 
IFRNet contains an auxiliary classifier that receives the latent representations of the filtered images as the input, and predicts their filter types. These predictions provide the extra information to the discriminators about the input images, and thus it makes training of discriminators more stabilized. Note that employing an auxiliary classifier does not improve the performance significantly. The general formula of our adversarial training can be seen in Equation \ref{eq:adv}.

\captionsetup[subfigure]{labelfont=bf, labelformat=empty}
\begin{figure*}[!ht]
        \centering
        \begin{subfigure}{0.138\textwidth}
                \includegraphics[width=\textwidth]{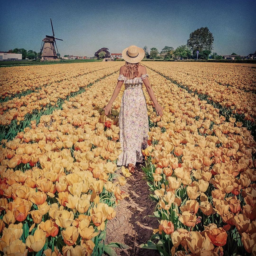}
                \includegraphics[width=\textwidth]{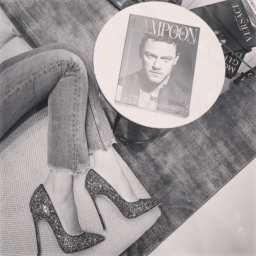}
                \includegraphics[width=\textwidth]{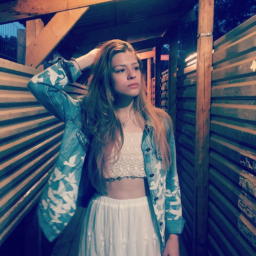}
                \includegraphics[width=\textwidth]{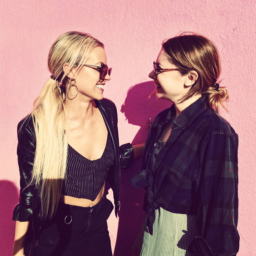}
                \includegraphics[width=\textwidth]{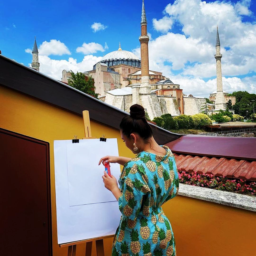}
                \includegraphics[width=\textwidth]{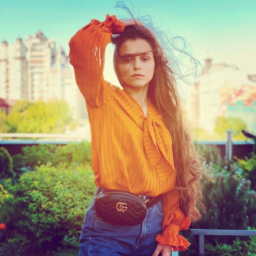}
                \caption{Filtered}
                \label{fig:fig4-filtered}
        \end{subfigure}       
        \begin{subfigure}{0.138\textwidth}
                \includegraphics[width=\textwidth]{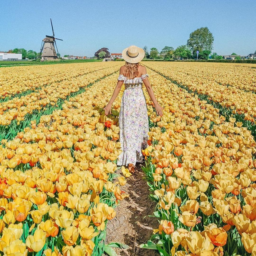}
                \includegraphics[width=\textwidth]{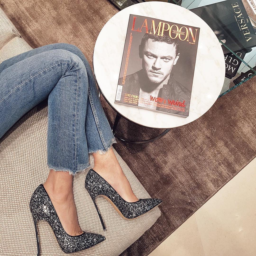}
                \includegraphics[width=\textwidth]{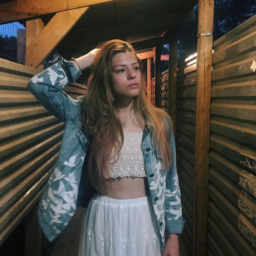}
                \includegraphics[width=\textwidth]{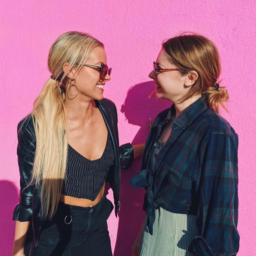}
                \includegraphics[width=\textwidth]{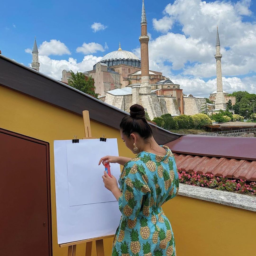}
                \includegraphics[width=\textwidth]{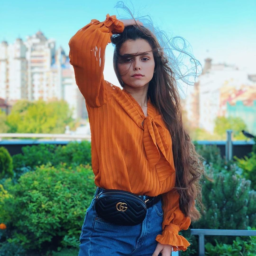}
                \caption{Original}
                \label{fig:fig4-org}
        \end{subfigure}
        \begin{subfigure}{0.138\textwidth}
                \includegraphics[width=\textwidth]{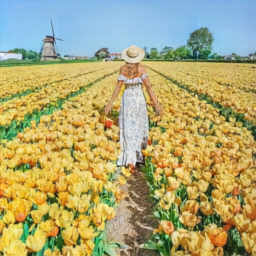}
                \includegraphics[width=\textwidth]{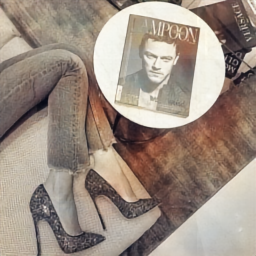}
                \includegraphics[width=\textwidth]{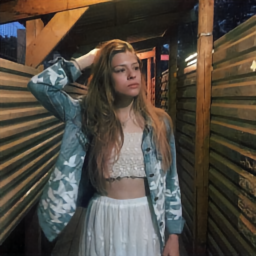}
                \includegraphics[width=\textwidth]{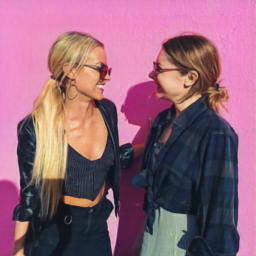}
                \includegraphics[width=\textwidth]{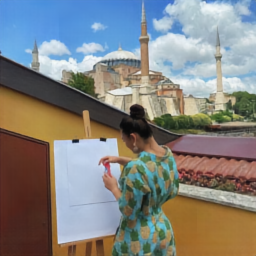}
                \includegraphics[width=\textwidth]{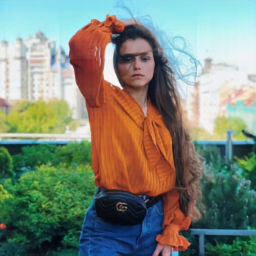}
                \caption{IFRNet \textbf{(ours)}}
                \label{fig:fig4-ifrnet}
        \end{subfigure}
        \begin{subfigure}{0.138\textwidth}
                \includegraphics[width=\textwidth]{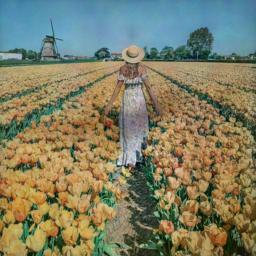}
                \includegraphics[width=\textwidth]{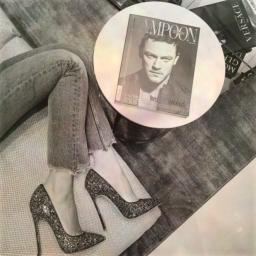}
                \includegraphics[width=\textwidth]{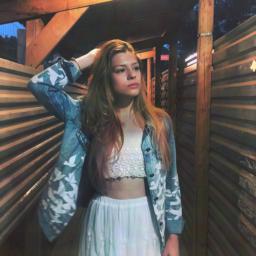}
                \includegraphics[width=\textwidth]{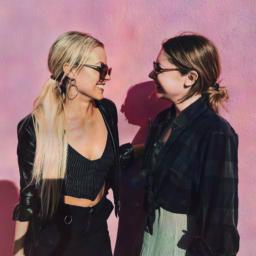}
                \includegraphics[width=\textwidth]{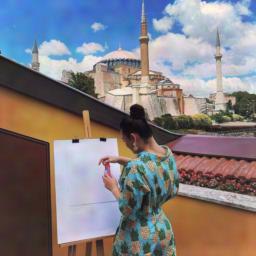}
                \includegraphics[width=\textwidth]{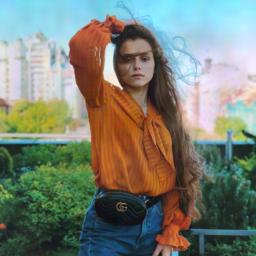}
                \caption{PE \cite{artisticFilter}}
                \label{fig:fig4-pe}
        \end{subfigure}
        \begin{subfigure}{0.138\textwidth}
                \includegraphics[width=\textwidth]{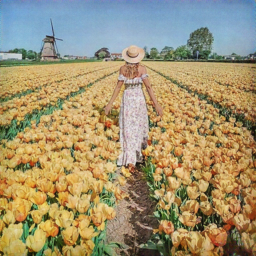}
                \includegraphics[width=\textwidth]{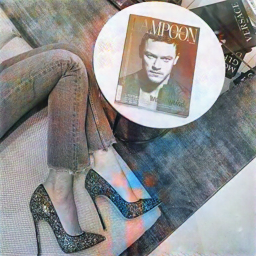}
                \includegraphics[width=\textwidth]{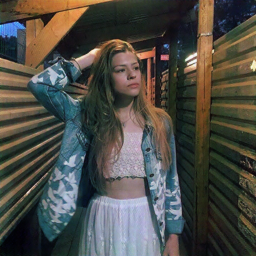}
                \includegraphics[width=\textwidth]{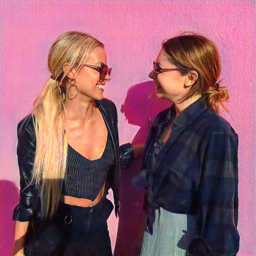}
                \includegraphics[width=\textwidth]{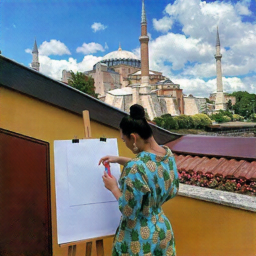}
                \includegraphics[width=\textwidth]{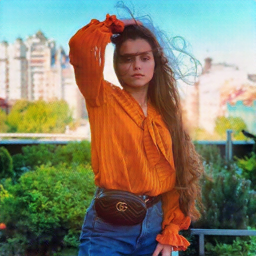}
                \caption{pix2pix \cite{pix2pix2017}}
                \label{fig:fig4-pix2pix}
        \end{subfigure}
        \begin{subfigure}{0.138\textwidth}
                \includegraphics[width=\textwidth]{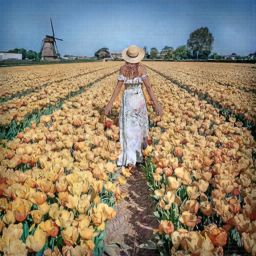}
                \includegraphics[width=\textwidth]{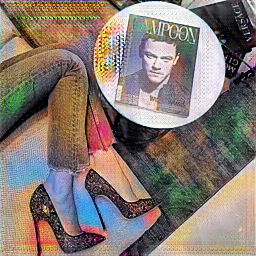}
                \includegraphics[width=\textwidth]{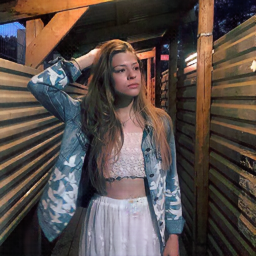}
                \includegraphics[width=\textwidth]{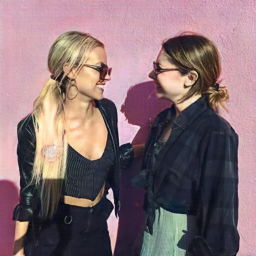}
                \includegraphics[width=\textwidth]{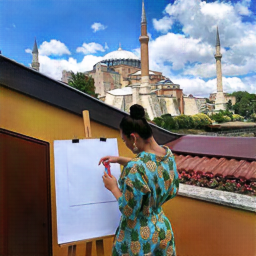}
                \includegraphics[width=\textwidth]{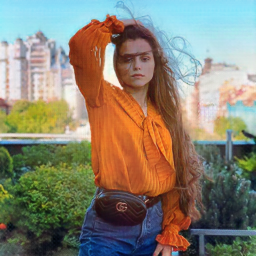}
                \caption{CycleGAN \cite{CycleGAN2017}}
                \label{fig:fig4-cycle}
        \end{subfigure}
        \begin{subfigure}{0.138\textwidth}
                \includegraphics[width=\textwidth]{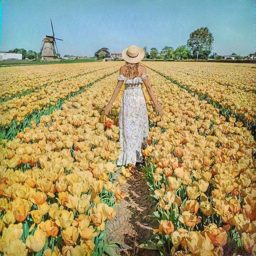}
                \includegraphics[width=\textwidth]{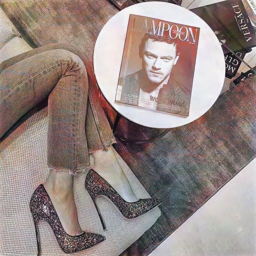}
                \includegraphics[width=\textwidth]{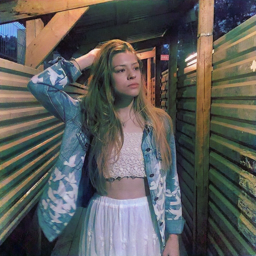}
                \includegraphics[width=\textwidth]{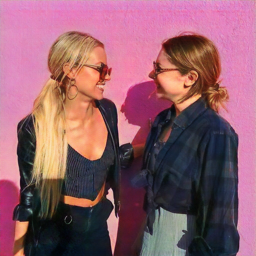}
                \includegraphics[width=\textwidth]{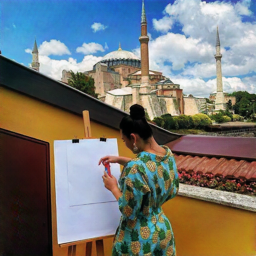}
                \includegraphics[width=\textwidth]{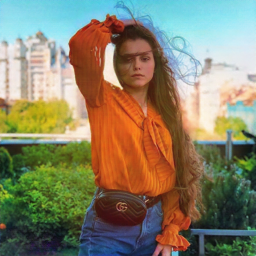}
                \caption{AngularGAN \cite{sidorov2019conditional}}
                \label{fig:fig4-angular}
        \end{subfigure}

        \caption{Comparison of the qualitative results of Instagram filter removal on IFFI dataset. Filters applied (top to bottom): \textit{Sutro}, \textit{Willow}, \textit{Nashville}, \textit{Amaro}, \textit{Lo-Fi}, \textit{Toaster}.}\label{fig:fig4} 
\end{figure*}

\begin{equation}
    \mathcal{L}_{adv}=\mathcal{L}_{glo}+\mathcal{L}_{loc}+\lambda_{gp}\mathcal{L}_{gp}+\lambda_{cls}\mathcal{L}_{cls}
    \label{eq:adv}
\end{equation}
where $\mathcal{L}_{glo}$ and $\mathcal{L}_{loc}$ represent the objective functions for global and local discriminators, respectively. $\mathcal{L}_{gp}$ is the gradient penalty whose weight $\lambda_{gp}$ is set to 10, and $\mathcal{L}_{cls}$ is the classification loss whose weight $\lambda_{cls}$ is set to 0.5. Finally, our objective function for IFRNet can be represented as follows:

\begin{equation}
    \mathcal{L} = \lambda_{tex}\mathcal{L}_{tex} + \lambda_{sem}\mathcal{L}_{sem} + \lambda_{adv}\mathcal{L}_{adv}
\end{equation}

\subsection{Experimental Setup}

In our study, we used our dataset, namely IFFI dataset, which contains 500 training and 100 test images combined with the set of their filtered versions with 16 different Instagram filters. We trained IFRNet with the resized images (\ie $256\times256$) for 120,000 steps with the batch size of 8. We only applied horizontal flipping to the images, no other data augmentation technique is applied. We used Adam optimizer \cite{KingmaB14} with $\beta_{1} = 0.5$ and $\beta_{2} =0.9$ for both generator and discriminator networks. The learning rate for generator network is $2 \times 10^{-4}$, and $10^{-3}$ for discriminator networks, and we did not change it during training. The weights of the main components of our objective function can be seen as follows: $\lambda_{tex}=10^{-3}$, $\lambda_{sem}=10^{-4}$ and $\lambda_{adv}=10^{-3}$. The implementation of IFRNet is done on PyTorch \cite{NEURIPS2019_9015}. The experiments have been conducted on 2x NVIDIA RTX 2080Ti GPUs, and a single run takes approximately 2 days to be completed. The source code can be found at \url{https://github.com/birdortyedi/instagram-filter-removal-pytorch}. 

\section{Results}

In this study, we compare the performance of IFRNet against a recent approach using CNNs for removing Instagram filters \cite{artisticFilter}, and also well-known paired/unpaired image-to-image translation studies including pix2pix \cite{pix2pix2017}, CycleGAN \cite{CycleGAN2017} and AngularGAN \cite{sidorov2019conditional}. For all these different methods, we have conducted several experiments on IFFI dataset with their default training settings.

\subsection{Qualitative Comparison}
\label{sec:qual}

As shown in Figure \ref{fig:fig4}, our proposed IFRNet achieves superior performance on Instagram filter removal when compared to the other methods. IFRNet can remove the visual effects injected to the images by filters at a large scale, while the other methods produce the outputs with significant differences from their original versions (\ie inconsistent background or foreground colors, some artifacts on the corners and the residuals of the filters). Therefore, we demonstrate that the external style information can be swept away from the images by normalizing the feature maps at each level in the encoder.

It is important to remark that some filters (\eg \textit{Toaster}, \textit{Sutro}, \textit{Willow}, \textit{Brannan}) apply several different transformations to the images, which may substantially alter the important details in the images. For example, \textit{Toaster} adds vignette and burning effects to the image (see the last row in Figure \ref{fig:fig4}), or \textit{Willow} directly attacks to the color information, and transform the image into a purplish gray-like image (see the second row in Figure \ref{fig:fig4}). Although the compared methods struggle to recover the images filtered by such challenging filters, IFRNet is able to remove these filters within a certain extent. Note that IFRNet does not have an objective function specialized to colorize the images, as in \cite{zhang2017real, zhang2016colorful}, hence it cannot recover the images filtered by \textit{Willow} as well as the other filters. More examples recovered by IFRNet are shown in Figure \ref{fig:fig7}.

\subsection{Quantitative Analysis}
\label{sec:quan}

We have employed four common image similarity metrics in our experiments to evaluate the quantitative performance of IFRNet, they can be specified as SSIM, PSNR, Learned Perceptual Image Patch Similarity (LPIPS) \cite{zhang2018perceptual} and CIE 2000 Color Difference (CIE-$\Delta$E) \cite{cie2000}. 

Table \ref{tab:tab-1} summarizes the performances of our proposed model and the other compared methods trained on IFFI dataset. Our proposed model outperforms the other compared methods on all common metrics. Particularly, the previous filter removal approach using CNNs \cite{artisticFilter} has limited performance, especially on the metrics prioritizing to measure the structural and perceptual similarity between the images (\ie SSIM and LPIPS), since it does not take any advantage of adversarial training. \textit{CycleGAN} \cite{CycleGAN2017} suffers from its inherent design, which is suitable for unpaired many-to-many image translation tasks. Although this approach shows outstanding performance on numerous applications in different domains, it falls behind our approach, and also \textit{AngularGAN} \cite{sidorov2019conditional} (its following study specialized for color constancy) and even \textit{pix2pix} \cite{pix2pix2017} (its primary study proposing a general solution for the image-to-image translation tasks). This intrinsically indicates that the approach learning to translate the input image to the target domain with the help of cycle consistency loss may not work as well as the other methods in the case of the existence of the paired images for many-to-one translation tasks. Note that we have used the official implementations of the compared studies with their default hyper-parameter settings.

\begin{table}
\begin{center}
\resizebox{\linewidth}{!}{\begin{tabular}{|c|c|c|c|c|}
\hline
\textbf{Method} & \textbf{SSIM} $\uparrow$& \textbf{PSNR} $\uparrow$& \textbf{LPIPS} $\downarrow$& \textbf{CIE-$\Delta$E} $\downarrow$ \\
\hline\hline
PE \cite{artisticFilter} & 0.748 & 23.41 & 0.069 & 39.55 \\
pix2pix \cite{pix2pix2017} & 0.825 & 26.35 & 0.048 & 30.32 \\
CycleGAN \cite{CycleGAN2017} & 0.819 & 22.94 & 0.065 & 36.59 \\
AngularGAN \cite{sidorov2019conditional} & 0.846 & 26.30 & 0.048 & 31.11 \\
IFRNet \textbf{(ours)} & \textbf{0.864} & \textbf{30.46} & \textbf{0.025} & \textbf{20.72} \\
\hline
\end{tabular}}
\end{center}
\caption{Quantitative performance of our proposed model and other compared methods on IFFI dataset.}
\label{tab:tab-1}
\end{table}

\begin{figure}[!b]
    \centering
    \resizebox{\linewidth}{!}{
    \includegraphics[width=\textwidth]{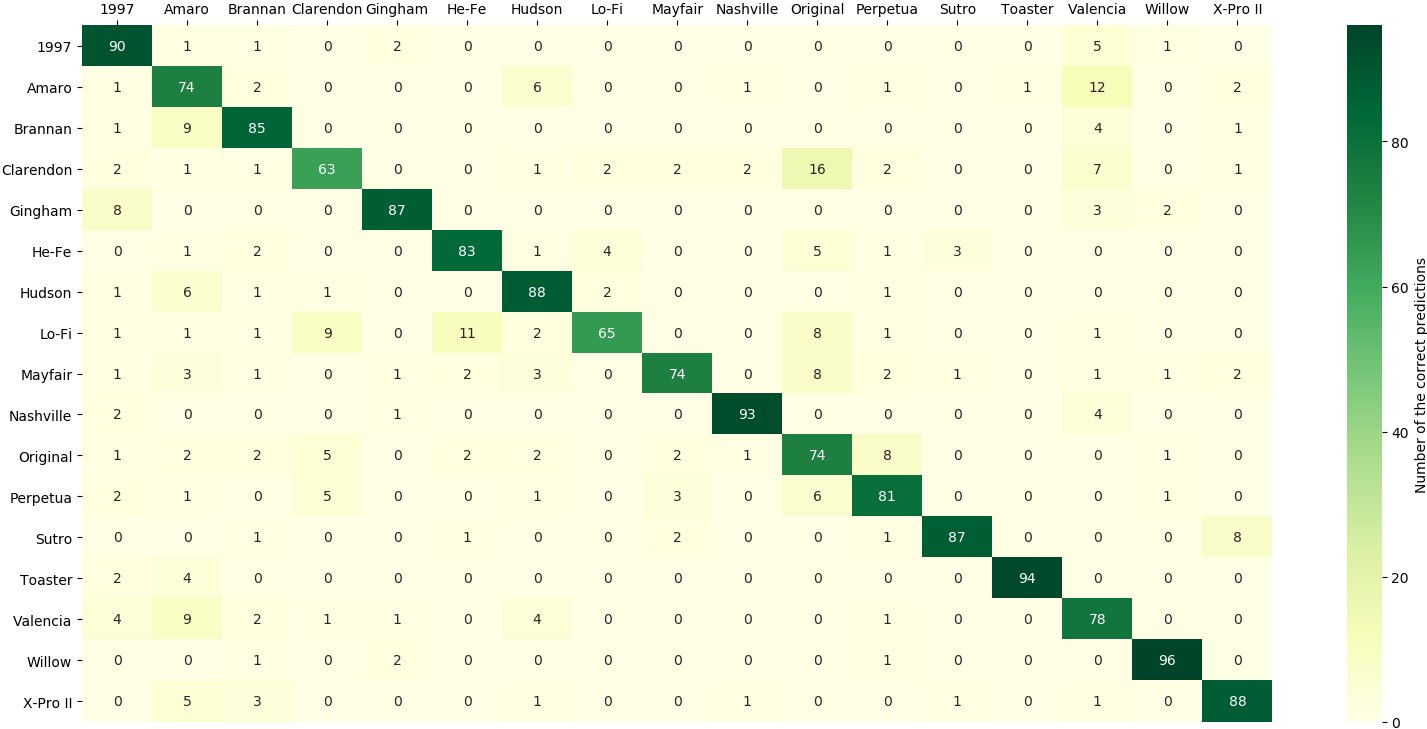}
    }
    
    \caption{Confusion matrix for the number of the correct predictions of auxiliary classifier of IFRNet for each filter on IFFI dataset.}\label{fig:fig5} 
\end{figure}

\subsection{Filter Classification}

In addition to removing Instagram filters from the images, we also demonstrate the performance of the auxiliary classifier of IFRNet on the same experimental setup. As we discuss in Section \ref{sec:ifr}, IFRNet has an auxiliary classifier that provides the extra information from the images to the discriminators to stabilize the adversarial training. This branch basically gives an output for the filter types of the images. 

IFRNet achieves 87.5\% overall classification accuracy on IFFI dataset. Considering classifying each filter as a separate task, we have measured the classification performance of IFRNet on each filter in order to evaluate how hard to differentiate these filters. As shown in Figure \ref{fig:fig5}, our proposed model does not perform the same for all filters, and gives better results for the filters containing different remarkable transformations (\eg \textit{Sutro}, \textit{Willow} and \textit{Toaster}). As aforementioned in Section \ref{sec:qual}, these particular filters dramatically change the visual details in the images, and thus it makes the process of classification of these filters less complicated. On the contrary, IFRNet gets more confused on differentiating more perceptibly similar filters, like \textit{Amaro}-\textit{Valencia}, \textit{He-Fe}-\textit{Lo-Fi} and \textit{Clarendon}-\textit{Original}. Therefore, we can say that the filters containing several different transformations are more likely to be classified correctly, but it is harder to remove them from the images.

\subsection{Dominant Color Estimation Analysis}

Considering the practical applications of Instagram filter removal for visual understanding tasks, we analyze the performance of a simple dominant color estimation algorithm on recovered images by IFRNet and the other methods. The algorithm applies K-Means Clustering to the color space of the images, and finds $N$ cluster centers (\ie the dominant colors). For our proposed model and the other methods, we compare CIE 2000 Color Difference (CIE-$\Delta$E) \cite{cie2000} of all dominant colors extracted by this algorithm. 

\begin{figure}[!t]
        \centering
        \begin{subfigure}{0.148\textwidth}
        \caption{}
        \includegraphics[width=\textwidth]{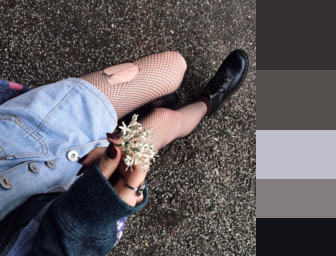}
        \caption{Valencia}
        \includegraphics[width=\textwidth]{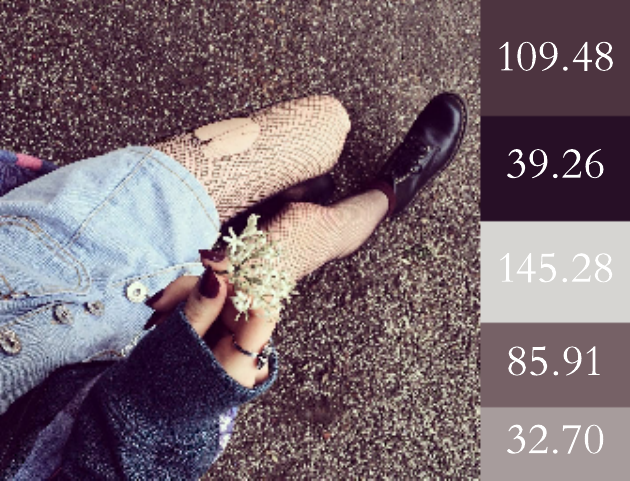}
        \caption{}
        \includegraphics[width=\textwidth]{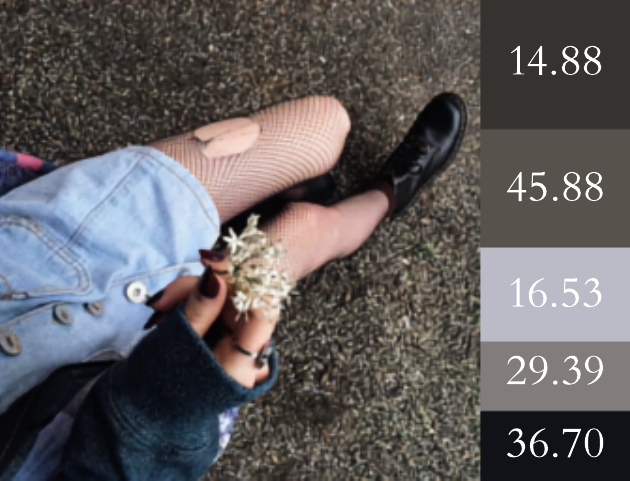}
        \caption{}
        \includegraphics[width=\textwidth]{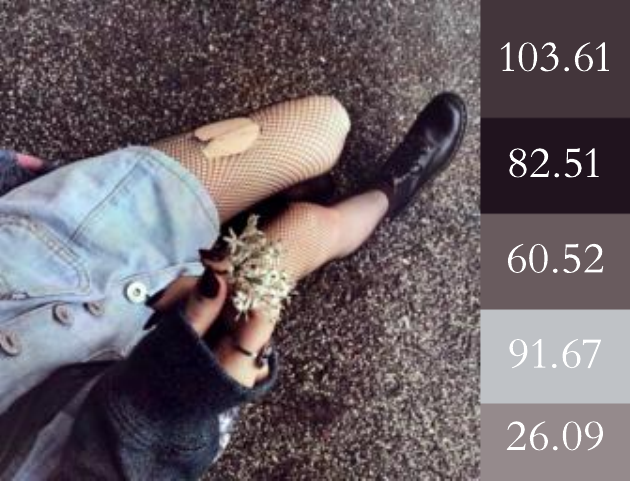}
        \caption{}
        \includegraphics[width=\textwidth]{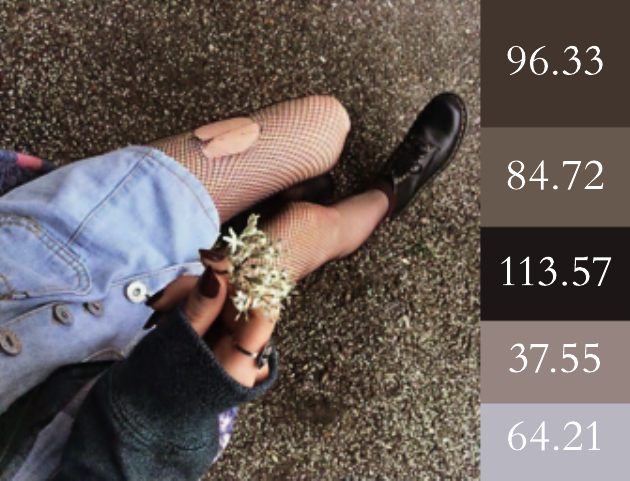}
        \caption{}
        \includegraphics[width=\textwidth]{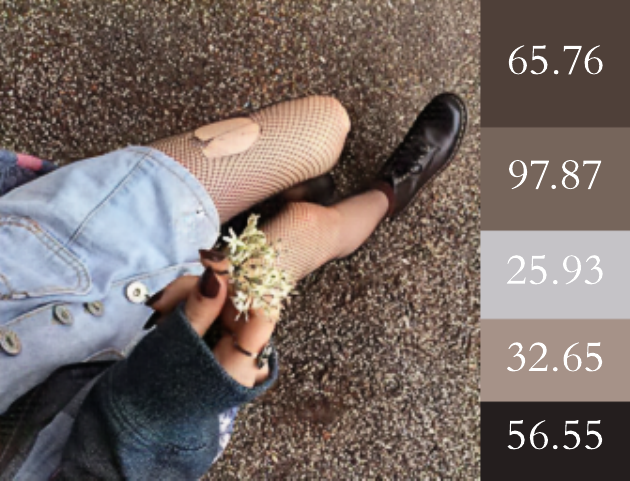}
        \caption{}
        \includegraphics[width=\textwidth]{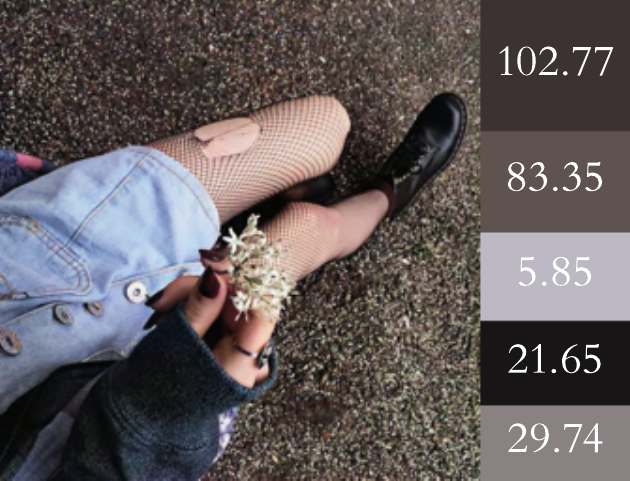}
        \end{subfigure} \hfill      
        \begin{subfigure}{0.148\textwidth}
        \caption{Original}
        \includegraphics[width=\textwidth]{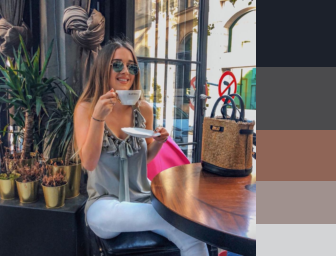}
        \caption{Sutro}
        \includegraphics[width=\textwidth]{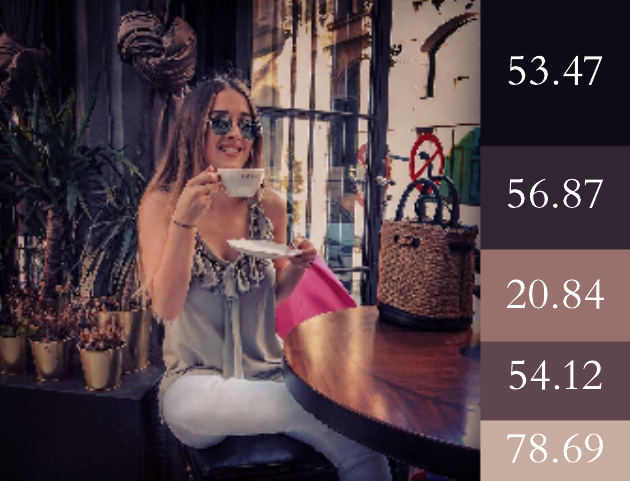}
        \caption{IFRNet \textbf{(ours)}}
        \includegraphics[width=\textwidth]{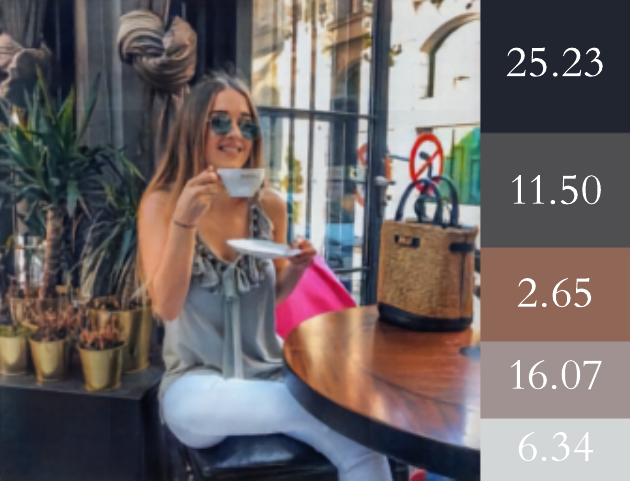}
        \caption{PE \cite{artisticFilter}}
        \includegraphics[width=\textwidth]{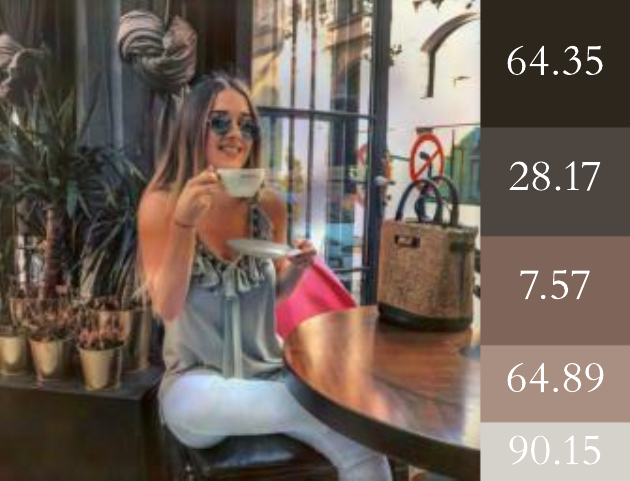}
        \caption{pix2pix \cite{pix2pix2017}}
        \includegraphics[width=\textwidth]{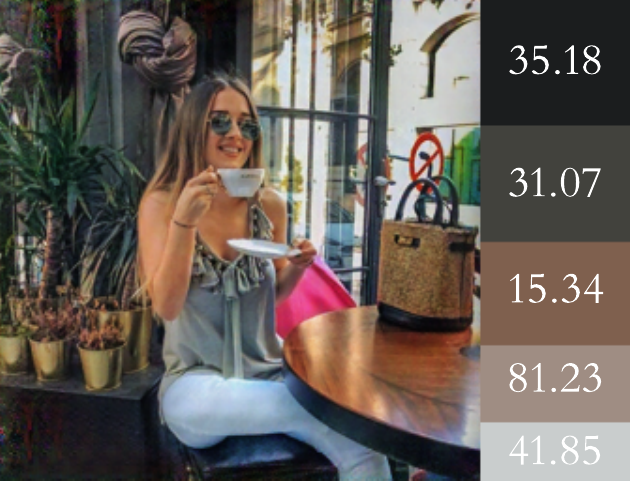}
        \caption{CycleGAN \cite{CycleGAN2017}}
        \includegraphics[width=\textwidth]{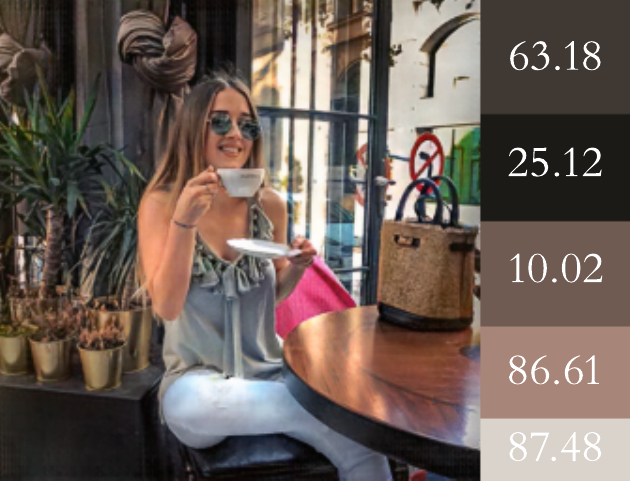}
        \caption{AngularGAN \cite{sidorov2019conditional}}
        \includegraphics[width=\textwidth]{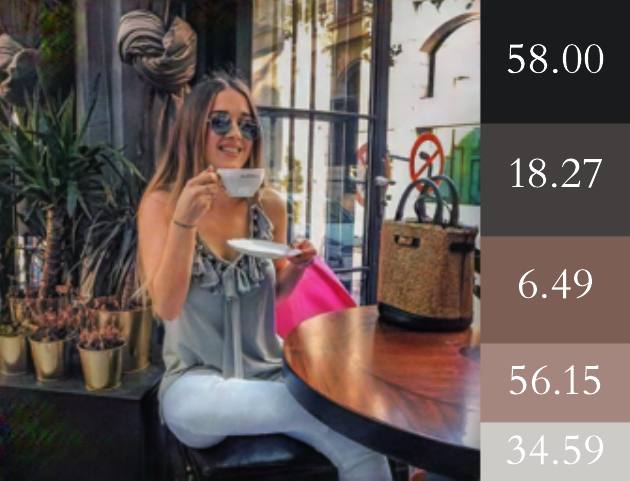}
        \end{subfigure} \hfill
        \begin{subfigure}{0.148\textwidth}
        \caption{}
        \includegraphics[width=\textwidth]{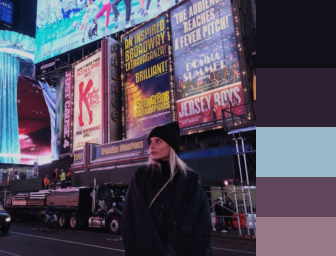}
        \caption{Toaster}
        \includegraphics[width=\textwidth]{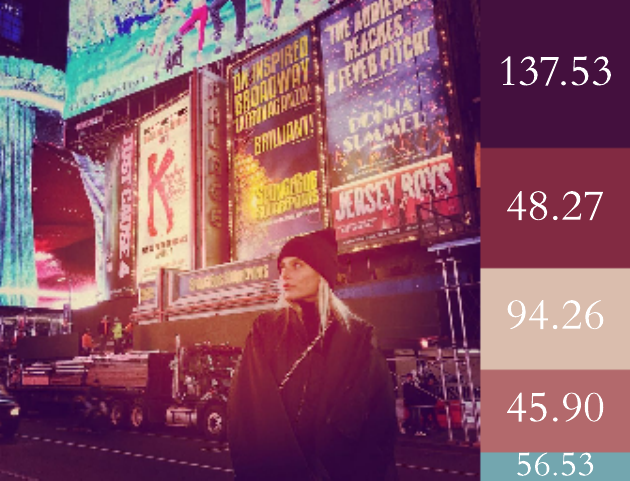}
        \caption{}
        \includegraphics[width=\textwidth]{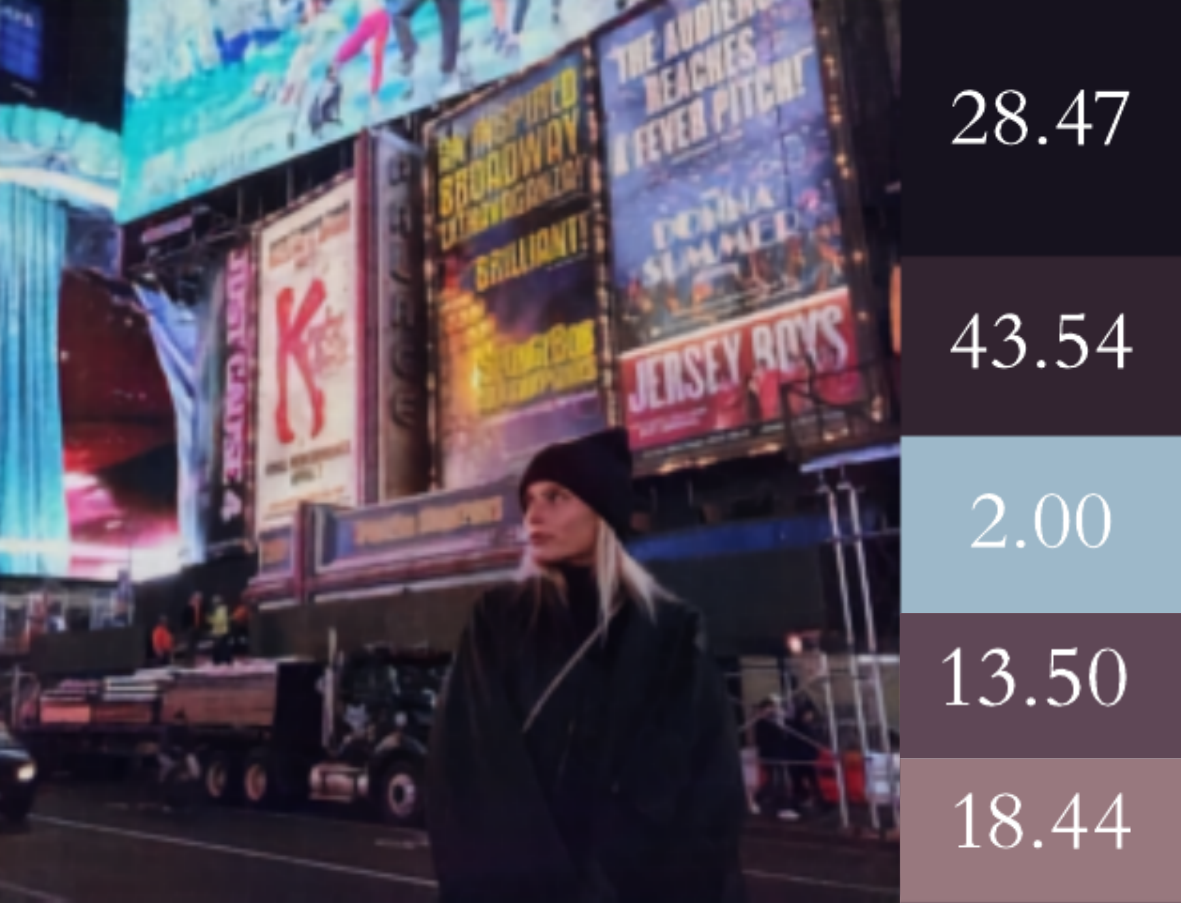}
        \caption{}
        \includegraphics[width=\textwidth]{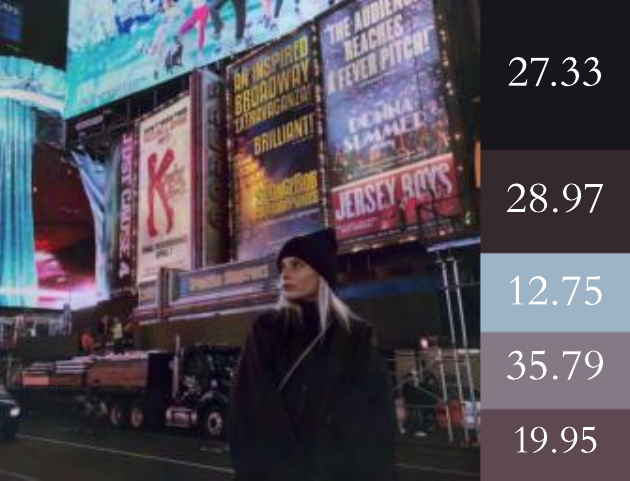}
        \caption{}
        \includegraphics[width=\textwidth]{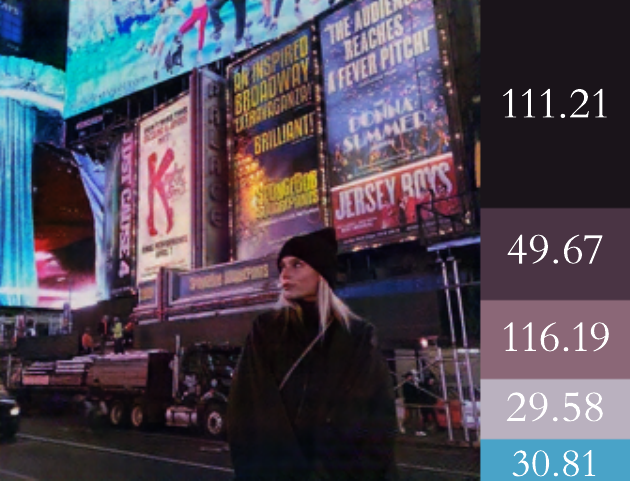}
        \caption{}
        \includegraphics[width=\textwidth]{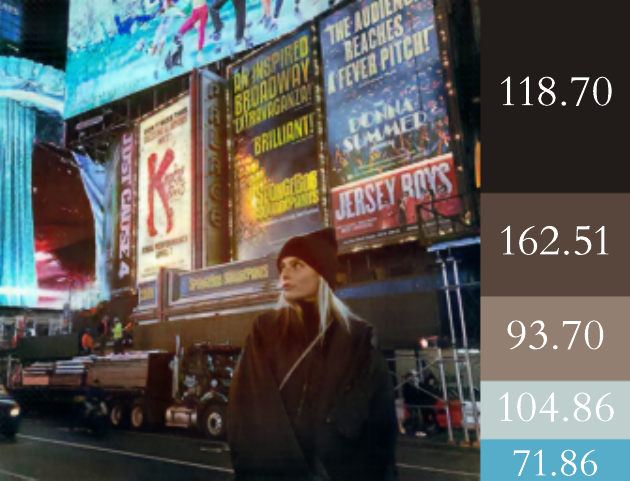}
        \caption{}
        \includegraphics[width=\textwidth]{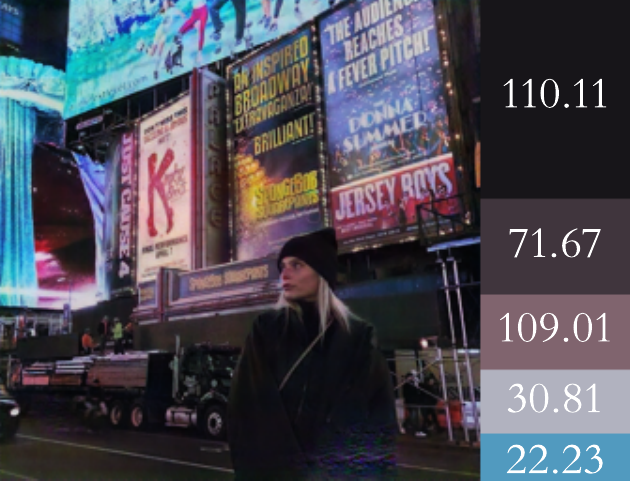}
        \end{subfigure}
        \caption{Example results of dominant color estimation on the images unfiltered by the compared methods. The bar on the right side of the images represents the weights of 5 dominant colors extracted by K-Means Clustering, and color distances (CIE 2000 \cite{cie2000}) to its corresponding color in the original image are shown on each color of this bar.}\label{fig:fig6} 
\end{figure}

\begin{figure*}[!t]
        \centering
        \begin{subfigure}{0.138\textwidth}
        \includegraphics[width=\textwidth]{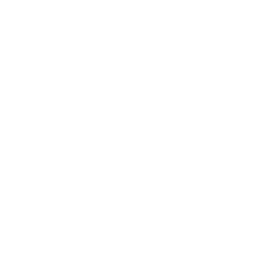}
        \includegraphics[width=\textwidth]{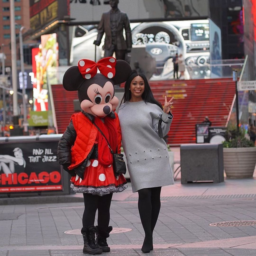}
        \includegraphics[width=\textwidth]{imgs/fig6/empty.png}
        \includegraphics[width=\textwidth]{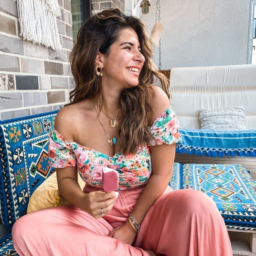}
        \includegraphics[width=\textwidth]{imgs/fig6/empty.png}
        \includegraphics[width=\textwidth]{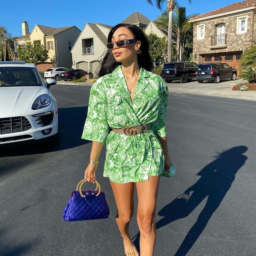}
        \includegraphics[width=\textwidth]{imgs/fig6/empty.png}
        \includegraphics[width=\textwidth]{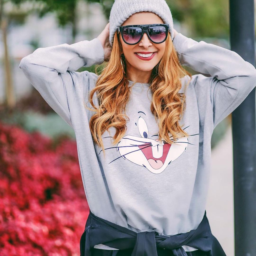}
        \label{fig:fig7-1}
        \end{subfigure}       
        \begin{subfigure}{0.138\textwidth}
        \includegraphics[width=\textwidth]{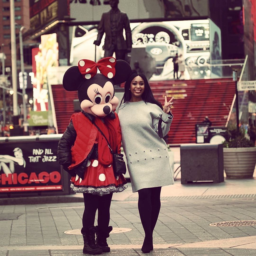}
        \includegraphics[width=\textwidth]{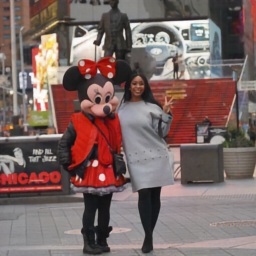}
        \includegraphics[width=\textwidth]{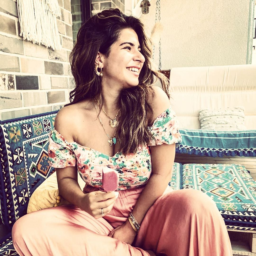}
        \includegraphics[width=\textwidth]{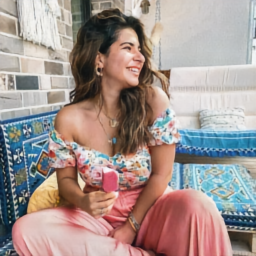}
        \includegraphics[width=\textwidth]{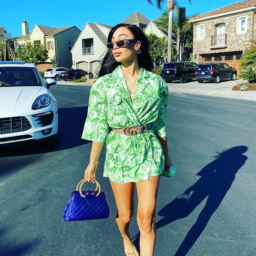}
        \includegraphics[width=\textwidth]{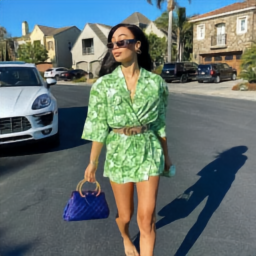}
        \includegraphics[width=\textwidth]{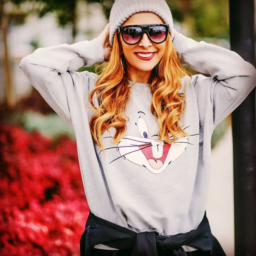}
        \includegraphics[width=\textwidth]{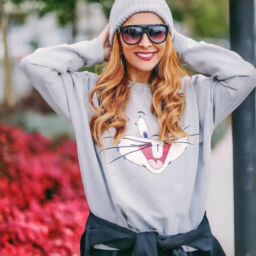}
        \label{fig:fig7-2}
        \end{subfigure}
        \begin{subfigure}{0.138\textwidth} \includegraphics[width=\textwidth]{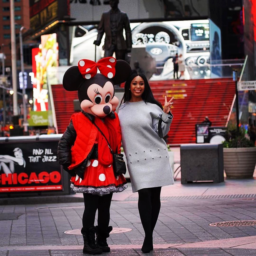}
        \includegraphics[width=\textwidth]{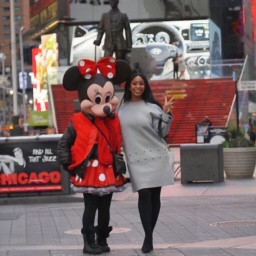}
        \includegraphics[width=\textwidth]{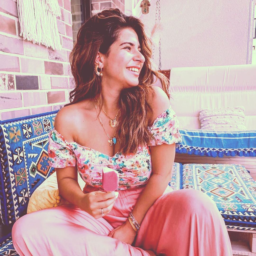}
        \includegraphics[width=\textwidth]{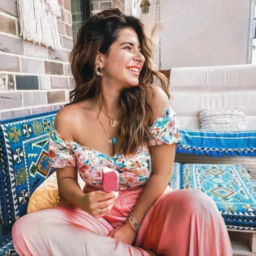}
        \includegraphics[width=\textwidth]{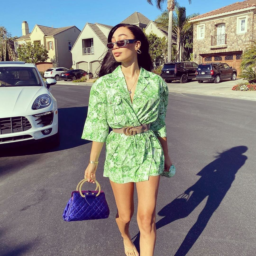}
        \includegraphics[width=\textwidth]{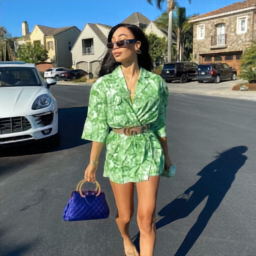}
        \includegraphics[width=\textwidth]{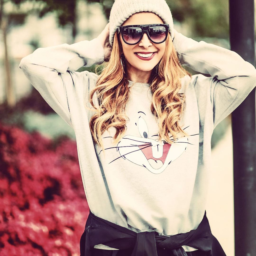}
        \includegraphics[width=\textwidth]{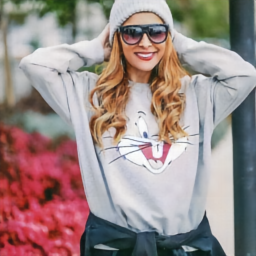}
        \label{fig:fig7-3}
        \end{subfigure}
        \begin{subfigure}{0.138\textwidth} \includegraphics[width=\textwidth]{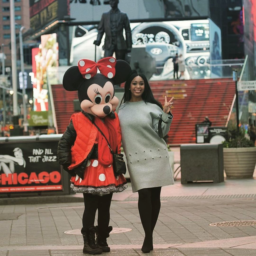}
        \includegraphics[width=\textwidth]{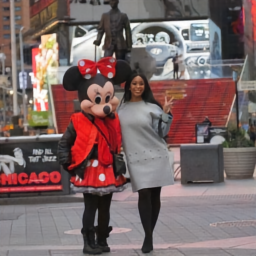}
        \includegraphics[width=\textwidth]{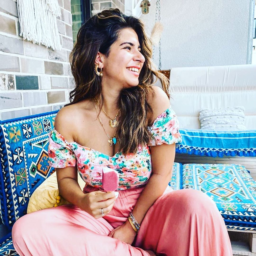}
        \includegraphics[width=\textwidth]{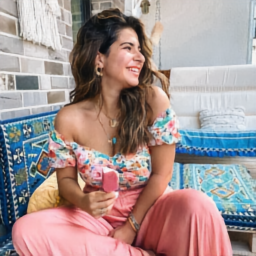}
        \includegraphics[width=\textwidth]{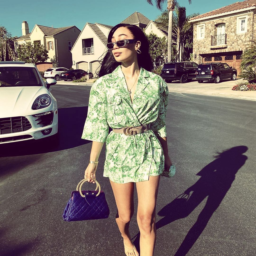}
        \includegraphics[width=\textwidth]{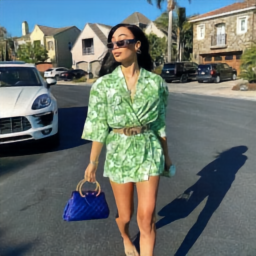}
        \includegraphics[width=\textwidth]{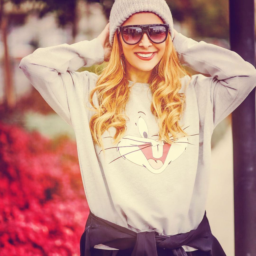}
        \includegraphics[width=\textwidth]{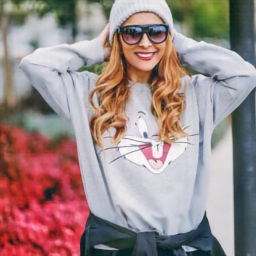}
        \label{fig:fig7-4}
        \end{subfigure}
        \begin{subfigure}{0.138\textwidth} \includegraphics[width=\textwidth]{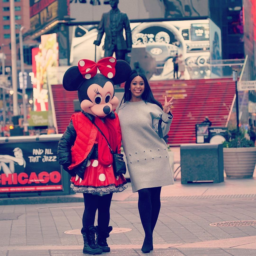}
        \includegraphics[width=\textwidth]{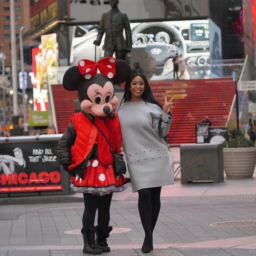}
        \includegraphics[width=\textwidth]{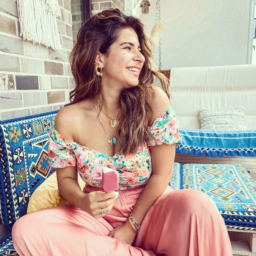}
        \includegraphics[width=\textwidth]{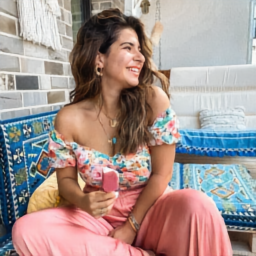}
        \includegraphics[width=\textwidth]{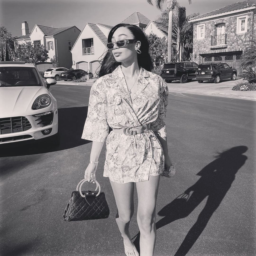}
        \includegraphics[width=\textwidth]{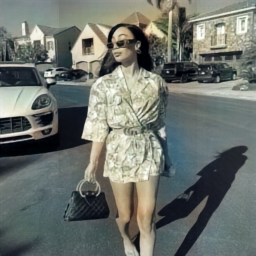}
        \includegraphics[width=\textwidth]{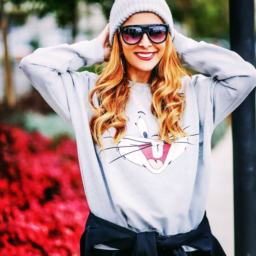}
        \includegraphics[width=\textwidth]{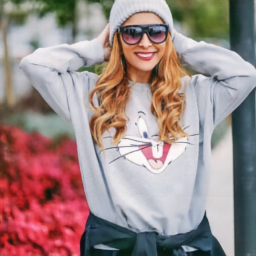}
        \label{fig:fig7-5}
        \end{subfigure}
        \begin{subfigure}{0.138\textwidth}
            \includegraphics[width=\textwidth]{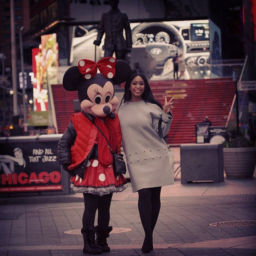}
            \includegraphics[width=\textwidth]{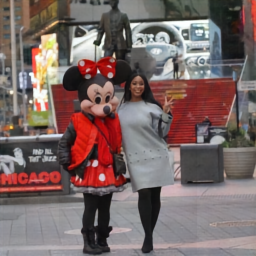}
            \includegraphics[width=\textwidth]{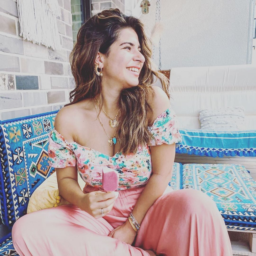}
            \includegraphics[width=\textwidth]{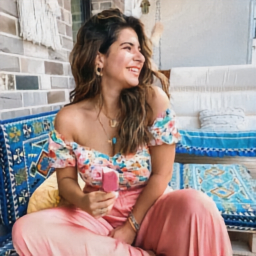}
            \includegraphics[width=\textwidth]{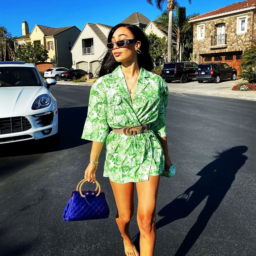}
            \includegraphics[width=\textwidth]{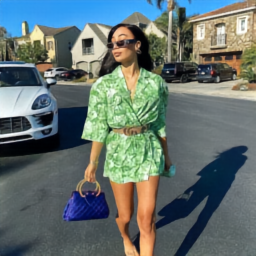}
            \includegraphics[width=\textwidth]{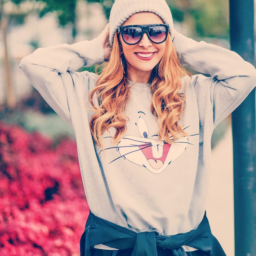}
            \includegraphics[width=\textwidth]{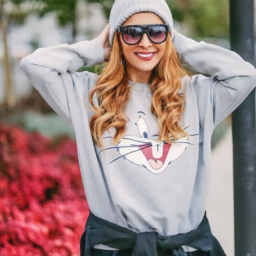}
            \label{fig:fig7-6}
        \end{subfigure}
        \begin{subfigure}{0.138\textwidth} 
        \includegraphics[width=\textwidth]{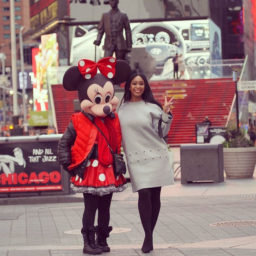}
        \includegraphics[width=\textwidth]{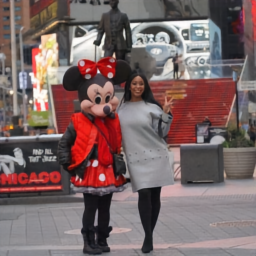}
        \includegraphics[width=\textwidth]{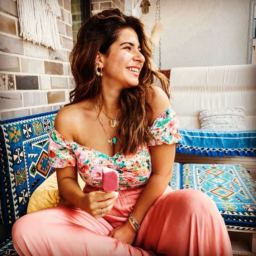}
        \includegraphics[width=\textwidth]{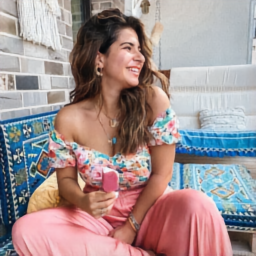}
        \includegraphics[width=\textwidth]{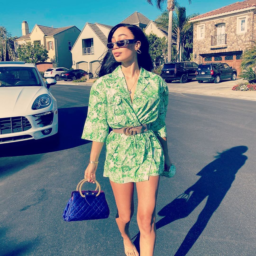}
        \includegraphics[width=\textwidth]{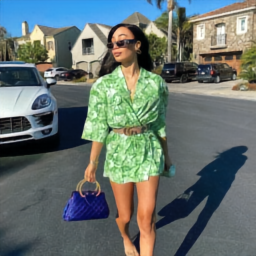}
        \includegraphics[width=\textwidth]{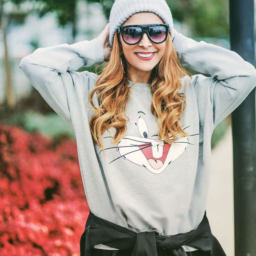}
        \includegraphics[width=\textwidth]{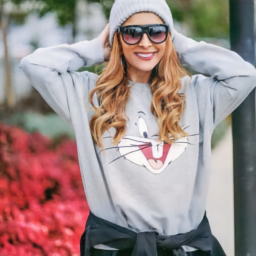}
        \label{fig:fig7-7}
        \end{subfigure}
        \caption{More examples from the results of IFRNet on IFFI dataset. Rows: (Odd) Filtered images, (Even) Original images at the first column, the rest represents the recovered versions of corresponding filtered images. The filters applied (left to right) for the first image: \textit{Brannan}, \textit{Lo-Fi}, \textit{Perpetua}, \textit{Nashville}, \textit{Sutro}, \textit{Valencia}, for the second image: \textit{Brannan}, \textit{1977}, \textit{Clarendon}, \textit{Valencia}, \textit{Gingham}, \textit{He-Fe}, for third image: \textit{Clarendon}, \textit{Valencia}, \textit{Brannan}, \textit{Willow}, \textit{Lo-Fi}, \textit{Nashville}, for fourth image: \textit{He-Fe}, \textit{Brannan}, \textit{Toaster}, \textit{Lo-Fi}, \textit{Nashville}, \textit{Perpetua}.}\label{fig:fig7} 
\end{figure*}

Example results of dominant color estimation algorithm on unfiltered images are shown in Figure \ref{fig:fig6}. Supporting the quantitative analysis in Section \ref{sec:quan}, our proposed model has the ability to recover the color information better than other compared methods, \textbf{even if it does not contain any special strategy for this information}. Note that IFRNet does not directly model the color conversion mapping, but learns the injected style information. Moreover, to compare the other methods, the images unfiltered by the previous study using CNNs by exploiting the polynomial expansion \cite{artisticFilter} and AngularGAN \cite{sidorov2019conditional} have closer and consistent dominant color predictions than the ones unfiltered by the prominent image-to-image translation methods \cite{pix2pix2017, CycleGAN2017}, since \cite{artisticFilter, sidorov2019conditional} contain different modules to specialize to gather color information.

\section{Conclusion}

In this study, we introduce IFRNet, an encoder-decoder structure applying adaptive feature normalization to all levels in the encoder to remove the external visual effects injected by filters. Experiments on Instagram filter removal task verify that IFRNet eliminates the external visual effects to a great extent. The main idea behind IFRNet is to consider the external visual effects as the style information, and we show that it is possible to remove them by normalizing the style of all feature maps, similar to the approach in style transfer studies. By extending the scope of the dataset for all available Instagram filters, this method could be employed for pre-processing the social media images before feeding them into a vision framework to enhance its performance. 

{\small
\bibliographystyle{ieee_fullname}
\bibliography{cvpr}
}

\end{document}